\documentclass[10pt,letterpaper]{technicalreport}
\usepackage{graphicx,booktabs,colortbl,multirow}
\usepackage{amsmath,amssymb}
\usepackage{algorithm,algpseudocode,wrapfig,needspace}
\usepackage{inconsolata}
\AtBeginEnvironment{verbatim}{\linespread{0.92}\selectfont}
\setcitestyle{citesep={;},aysep={,},yysep={;}}
\hypersetup{pdftitle={On the Importance of Hidden-State Refinement in Masked Diffusion Language Models},pdfauthor={Linhao Zhong, Bozhen Fang, Xuehao Wang, Zekai Luo, Liyang Li, Hourong Li, Chenchen Jing, Hao Chen}}

\newcommand{\finishfigurewrap}{%
  \par
  \ifnum\value{WF@wrappedlines}>1\relax
    \vskip\dimexpr\baselineskip*(\value{WF@wrappedlines}-1)\relax
  \fi
  \WFclear
}

\title{On the Importance of Hidden-State Refinement\\
in Masked Diffusion Language Models}
\author{Linhao~Zhong\textsuperscript{1,*},
Bozhen~Fang\textsuperscript{1,*},
Xuehao~Wang\textsuperscript{1},
Zekai~Luo\textsuperscript{1},\\
Liyang~Li\textsuperscript{1},
Hourong~Li\textsuperscript{1},
Chenchen~Jing\textsuperscript{2},
Hao~Chen\textsuperscript{1}}
\affiliations{\textsuperscript{1}Zhejiang University, State Lab of CAD \& CG\quad
\textsuperscript{2}Zhejiang University of Technology}
\contributions{\textsuperscript{*}Equal contribution.}
\projectpage{https://zhongzero.github.io/Recurrent-DLLM/}
\date{}

\begin{document}
\begin{reportfrontmatter}
\maketitle
\begin{abstract}
Masked diffusion language models (MDLMs) generate text by iteratively denoising masked or corrupted token sequences, allowing token predictions to condition on bidirectional context and update multiple positions in parallel. 
While each denoising step produces a rich layer-wise hidden-state trajectory, prior work typically focuses only on token-level information and discards these continuous hidden states, leaving their refinement dynamics underexplored. 
In this work, we analyze hidden-state dynamics in MDLMs and find clear evidence of layer-wise and cross-step refinement: smooth evolution across adjacent layers, greater predictive power in deeper layers, gradual degradation under transformer block removal, and strong final-state alignment across consecutive denoising steps. 
Motivated by these observations, we propose Cross-Step Recurrent Refinement (CSRR), a mechanism that carries hidden-state information across denoising steps to enhance generation. Extensive experiments show that CSRR improves text generation quality, highlighting the critical role of continuous hidden-state refinement in MDLMs.
\end{abstract}

\reportlinks
\end{reportfrontmatter}
\section{Introduction}

Masked diffusion language models (MDLMs)~\citep{mdlm,md4,llada,dream2025} have recently gained attention as an alternative to autoregressive models for text generation. Instead of relying on a strict left-to-right factorization, MDLMs start from a corrupted or masked token sequence and iteratively recover token identities through denoising steps~\citep{d3pm,diffusionbert,rdm}. This approach allows token predictions to condition on bidirectional context and update multiple positions simultaneously. Beyond updating masked tokens, each denoising step produces a full layer-wise trajectory of continuous hidden states,  which encode intermediate representations of the sequence before final token probabilities are computed.

\begin{figure}[htbp]
    \centering
    \includegraphics[width=\textwidth]{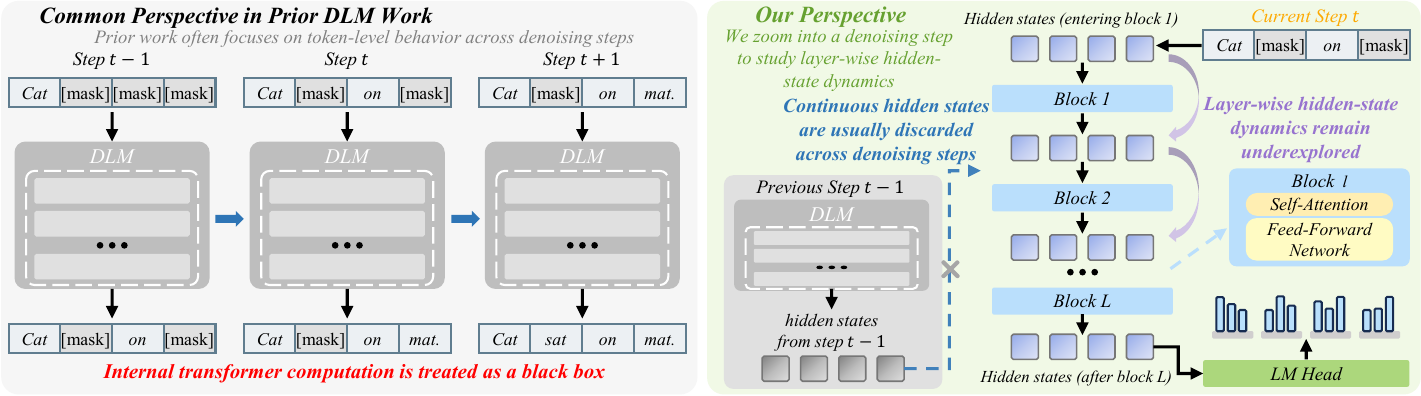}
    \caption{Perspective comparison for MDLM generation. Prior work commonly focuses on token-level behavior across denoising steps and treats the internal transformer computation as a black box. We instead zoom into the current denoising step $t$ to study layer-wise hidden-state dynamics, and ask whether hidden states from the previous step $t-1$ can be leveraged as cross-step refinement signals.}
    \label{fig:perspective-comparison}
\end{figure}

Prior work~\citep{mdlm,md4,llada,dream2025,wu2025fast,wu2025fast2} primarily focuses on token-level behavior across denoising steps, treating the internal transformer computation as a black box. Most sampling procedures~\citep{arriolablock,sdar,d2f,DAEDAL,liu2025sequential} also carry only token-level information between steps. As a result, the continuous hidden-state trajectory is discarded after each model call, and the next step mainly receives the updated token sequence. This gives rise to two limitations: (1) layer-wise hidden-state dynamics remain underexplored, leaving it unclear whether transformer layers simply transform representations or progressively refine them toward the final prediction; (2) discarding hidden states limits the ability of subsequent steps to leverage the rich continuous refinement information they contain (Figure~\ref{fig:perspective-comparison}).

Motivated by these limitations, we investigate whether continuous hidden states can be understood and leveraged as refinement signals in MDLM generation. Specifically, we address two research questions:

\begin{itemize}
    \item \textbf{RQ1.} Do hidden states in MDLMs exhibit layer-wise and cross-step refinement?
    \item \textbf{RQ2.} Can previous-step hidden states be leveraged to improve MDLM generation?
\end{itemize}

\noindent \textbf{Contribution 1.} To address RQ1, we conduct a diagnostic study of hidden-state refinement in MDLMs and highlight four key findings. First, hidden states evolve smoothly across adjacent layers, indicating incremental layer-wise refinement rather than abrupt transformations. Second, deeper layers progressively capture more predictive information for the final generated sequence, showing that refinement accumulates toward the output. Third, refinement is distributed across blocks and both self-attention and feedforward sublayers, with no single block dominating the computation. Fourth, final hidden states remain highly consistent across consecutive denoising steps, suggesting that previous-step states provide a reliable context for cross-step reuse. Together, these results demonstrate that MDLM generation involves rich, continuous layer-wise and cross-step refinement beyond token-level updates, highlighting a largely untapped source of information for improving generation.

\noindent \textbf{Contribution 2.} To answer RQ2, we propose Cross-Step Recurrent Refinement (CSRR), a lightweight recurrent mechanism that propagates hidden-state information from the previous denoising step to the current one. By injecting prior hidden states into the first block of the current step, CSRR allows the model to leverage accumulated refinement, enhancing the coherence and fidelity of generated sequences. We validate CSRR across multiple MDLM architectures and benchmark datasets, demonstrating consistent improvements in generation quality, convergence speed, and robustness. These results highlight the practical utility of continuous hidden-state refinement and suggest that incorporating such recurrent signals can be a general principle for improving iterative text generation models.
\section{Preliminaries for MDLMs}

\paragraph{Forward process.} Masked diffusion language models define a corruption process over token identities. Given a clean token sequence $x_0=(x_0^1,\dots,x_0^N)$, the forward process maps it to a partially masked sequence $\tilde{x}_\tau$ at noise level $\tau$. Let $\rho_\tau$ denote the masking rate at this level. A common forward process independently replaces each token with a special mask token $\mathtt{[MASK]}$:
\begin{equation}
q(\tilde{x}_\tau^i\mid x_0^i) =
\begin{cases}
1-\rho_\tau, & \tilde{x}_\tau^i=x_0^i, \\
\rho_\tau, & \tilde{x}_\tau^i=\mathtt{[MASK]}.
\end{cases}
\end{equation}

\paragraph{Reverse process.} During generation, let $x_s$ denote the current partially masked sequence at denoising step $s$. The reverse process uses a learned denoiser $p_\theta(x_0\mid x_s)$ to predict clean token identities from $x_s$. The model repeatedly predicts token distributions for currently masked positions, accepts a subset of token predictions according to the sampler, and keeps the remaining positions masked for later denoising steps. 

\paragraph{Layer-wise computation.} Within a single denoising step, the denoiser first embeds the current token sequence:
\begin{equation}
    H_s^0 = \mathrm{Embed}(x_s).
\end{equation}
It then applies a stack of $L$ pre-normalized transformer blocks. For block $\ell+1$ with $\ell=0,\dots,L-1$, we write the post-self-attention state as
\begin{equation}
    A_s^{\ell+1}
    =
    H_s^\ell
    +
    \mathrm{SA}_{\ell+1}\!\left(\mathcal{N}^{\mathrm{attn}}_{\ell+1}(H_s^\ell)\right),
\end{equation}
and the block output after the feed-forward network as
\begin{equation}
    H_s^{\ell+1}
    =
    A_s^{\ell+1}
    +
    \mathrm{FFN}_{\ell+1}\!\left(\mathcal{N}^{\mathrm{ffn}}_{\ell+1}(A_s^{\ell+1})\right).
\end{equation}
Here $\mathcal{N}$ denotes the normalization used by the implementation, and $\mathrm{FFN}$ denotes the block's feed-forward sublayer. After the last block, a final normalization and the language-model head produce token probabilities:
\begin{equation}
    Z_s = \mathcal{N}^{\mathrm{final}}(H_s^L),
\end{equation}
\begin{equation}
    p_\theta(x_0\mid x_s)
    =
    \mathrm{Softmax}\!\left(W_{\mathrm{lm}} Z_s\right).
\end{equation}
Standard sampling uses the final normalized representation to produce logits, applies token updates, and then recomputes a new hidden-state trajectory at the next denoising step.


\section{Understanding Hidden-State Refinement in MDLMs}
\label{sec:observations}

We first ask whether MDLM generation is iterative not only across denoising steps, but also inside a single model call. We inspect hidden-state trajectories, intermediate predictiveness, block-removal perturbations, and cross-step hidden-state consistency. Together, these diagnostics test whether intermediate hidden states behave like transient activations or like continuous refinement states.

\paragraph{Diagnostic setup.} Unless otherwise noted, diagnostics use LLaDA-8B-Instruct~\citep{llada} on the first 10 GSM8K~\citep{dataset-gsm8k} test examples, with generation length $128$, $64$ denoising steps, block length $32$, temperature $0$, and low-confidence remasking. Layer $0$ is the embedding output and layer $32$ is the final transformer layer. Detailed diagnostic definitions are provided in Appendix~\ref{sec:detailed-diagnostic-definitions}.

\subsection{Hidden States Evolve Smoothly Across Layers}
\label{sec:obs-smooth}

We measure whether representations change abruptly from layer to layer using mean token-wise cosine similarity:
\begin{equation}
    \mathrm{Sim}_{s,\ell}
    =
    \mathbb{E}_{i}
    \left[
    \cos\left(H_{s,i}^{\ell}, H_{s,i}^{\ell+1}\right)
    \right],
\end{equation}
where $i$ indexes token positions. We also compute full layer-pair similarity and relative update norm:
\begin{equation}
    r_{s,\ell}
    =
    \frac{\lVert H_s^{\ell+1} - H_s^\ell \rVert_2}
         {\lVert H_s^\ell \rVert_2}.
\end{equation}

\begin{figure}[htbp]
    \centering
    \includegraphics[width=\textwidth]{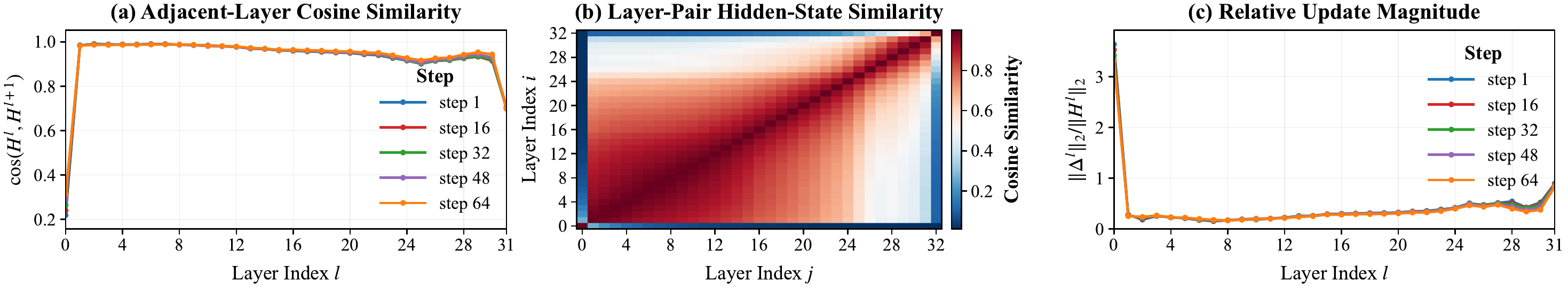}
    \caption{Hidden-state evolution across layers. The implemented figure combines adjacent-layer cosine similarity, layer-pair similarity, and relative update magnitude.}
    \label{fig:obs1}
\end{figure}

\paragraph{Adjacent-layer cosine similarity.} Figure~\ref{fig:obs1}(a) plots adjacent-layer cosine similarity for selected denoising steps. Across all 64 steps and 10 examples, the mean similarity is $0.930$, and middle layers are typically above $0.90$. Thus most transformer blocks update representations locally rather than replacing them abruptly, which is consistent with an incremental refinement process. The embedding-to-layer-1 transition is an exception, suggesting that the first block converts token embeddings into an initial contextual representation before later blocks perform incremental refinement. Appendix~\ref{sec:app-all-step-layer-smoothness} provides the all-step heatmap, and Appendix~\ref{sec:app-task-model-smoothness} shows the diagnostic across tasks and model variants.

\paragraph{Layer-pair hidden-state similarity.} Figure~\ref{fig:obs1}(b) extends the measurement to all layer pairs. The strong diagonal band shows that nearby layers remain close while distance grows gradually with depth, matching a continuous refinement trajectory rather than a sequence of unrelated transformations. Appendix~\ref{sec:app-cka-layer-similarity} repeats the analysis with linear CKA.

\paragraph{Relative update magnitude.} Figure~\ref{fig:obs1}(c) measures normalized update size between adjacent layers. High cosine similarity does not imply unchanged states, since each block still contributes a nonzero residual update. The result is therefore not that layers are idle, but that most layers make bounded corrections to an evolving hidden state. Together, the three panels support a smooth refinement view of layer-wise computation.

\subsection{Deeper Layers Become More Predictive}
\label{sec:obs-progressive}

Smoothness alone does not show that updates are useful. We therefore apply the model's final normalization and shared LM head to intermediate hidden states, and compare predictions with the final generated sequence used as the pseudo clean target. For layer $\ell$,
\begin{equation}
    Z_s^\ell = W_{\mathrm{lm}}\mathcal{N}^{\mathrm{final}}(H_s^\ell).
\end{equation}
After suppressing the mask-token logit, we compute token accuracy against the final generated sequence:
\begin{equation}
    \mathrm{Acc}_{s,\ell}
    =
    \mathrm{Acc}\left(\arg\max Z_s^\ell, x_{\mathrm{final}}\right).
\end{equation}

\begin{figure}[htbp]
    \centering
    \includegraphics[width=\textwidth]{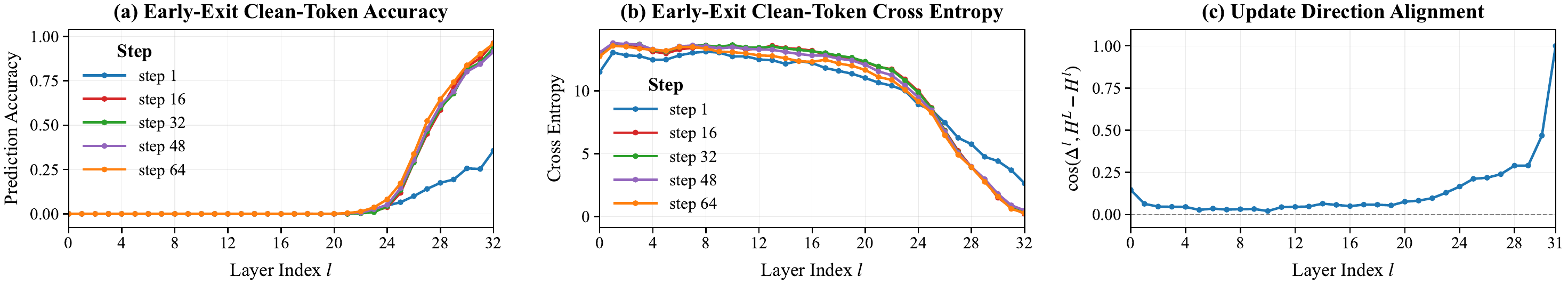}
    \caption{Deeper layers progressively improve denoising predictions. The implemented figure reports early-exit accuracy, cross entropy, and update-direction alignment.}
    \label{fig:obs2}
\end{figure}

\paragraph{Early-exit accuracy and cross entropy.} Panels (a) and (b) of Figure~\ref{fig:obs2} apply the shared LM head to intermediate layers. Early layers have little direct token-prediction ability, while late layers become much more predictive and have lower cross entropy. Thus the smooth trajectory is not merely geometrically smooth; it also accumulates information that becomes increasingly readable by the output head. Appendix~\ref{sec:app-topk-confidence} provides top-k accuracy, token-probability, and entropy diagnostics.

\begin{wrapfigure}{r}{0.5\textwidth}
    \centering
    \includegraphics[width=\linewidth]{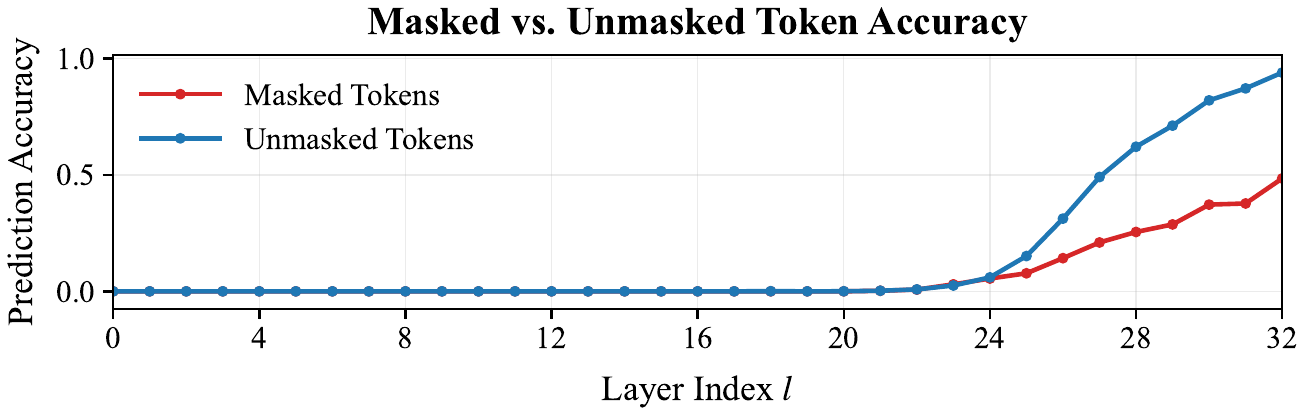}
    \caption{Masked-token and unmasked-token early-exit accuracy curves.}
    \label{fig:obs2-masked-unmasked}
\end{wrapfigure}

\paragraph{Update-direction alignment.} Figure~\ref{fig:obs2}(c) measures whether the local update $H^{\ell+1}-H^\ell$ points toward the final-layer residual direction $H^L-H^\ell$. Positive alignment indicates movement toward the final representation rather than orthogonal drift. This directional evidence complements the early-exit results and supports a view in which intermediate layers participate in denoising, rather than only preserving a stable representation. Appendix~\ref{sec:app-correction-break-rates} decomposes adjacent-layer prediction changes into wrong-to-correct fixes and correct-to-wrong breaks.

\paragraph{Masked vs. unmasked token accuracy.} Figure~\ref{fig:obs2-masked-unmasked} separates still-masked and already-unmasked positions. Both improve with depth, while unmasked positions remain easier because they contain more lexical evidence. The masked-token gains show that late layers recover genuinely hidden content rather than only preserving tokens that are already visible.

\begin{wrapfigure}{r}{0.5\textwidth}
    \centering
    \includegraphics[width=\linewidth]{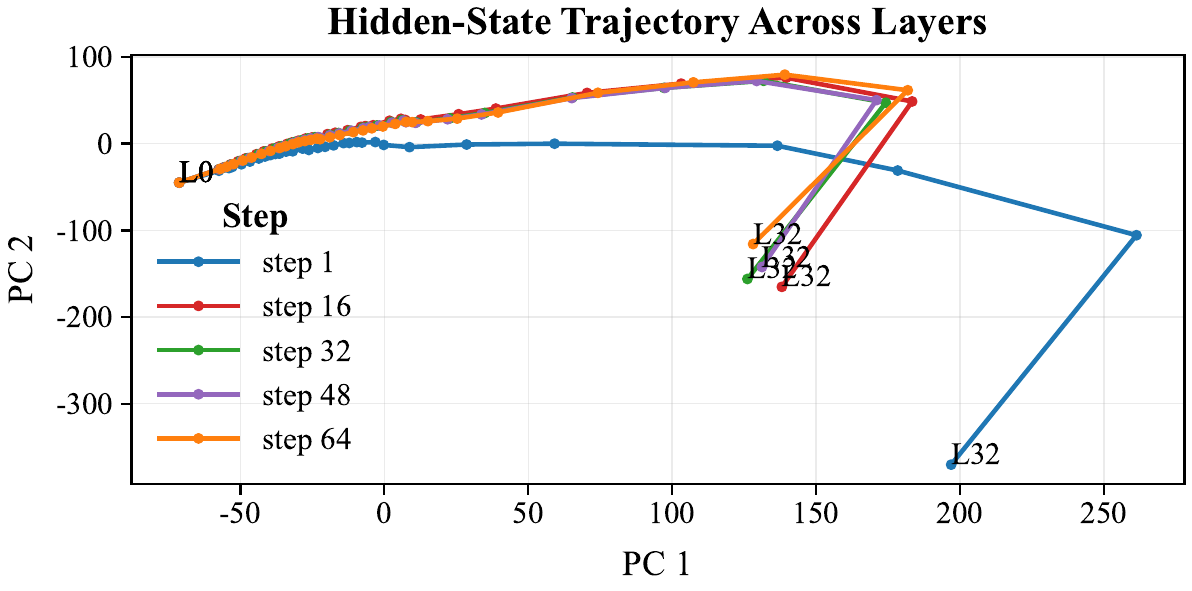}
    \caption{PCA trajectory of hidden states across layers.}
    \label{fig:obs2-trajectory}
\end{wrapfigure}

\paragraph{Hidden-state trajectory.} Figure~\ref{fig:obs2-trajectory} projects hidden states into a shared two-dimensional PCA space. The trajectory provides a qualitative companion to the accuracy and alignment diagnostics, showing that states move along a structured path across layers rather than scattering randomly. This visualization makes the layer-wise refinement interpretation more explicit.

\subsection{Refinement Is Distributed Across Blocks}
\label{sec:obs-distributed}

If denoising were concentrated in a single critical block, removing most other blocks would have little effect. We test this by replacing selected transformer blocks with the identity during inference, including single blocks, consecutive windows of $1$, $2$, $4$, or $8$ blocks, early/middle/late four-block regions, and self-attention-only or FFN-only sublayers.

\begin{figure}[htbp]
    \centering
    \includegraphics[width=\textwidth]{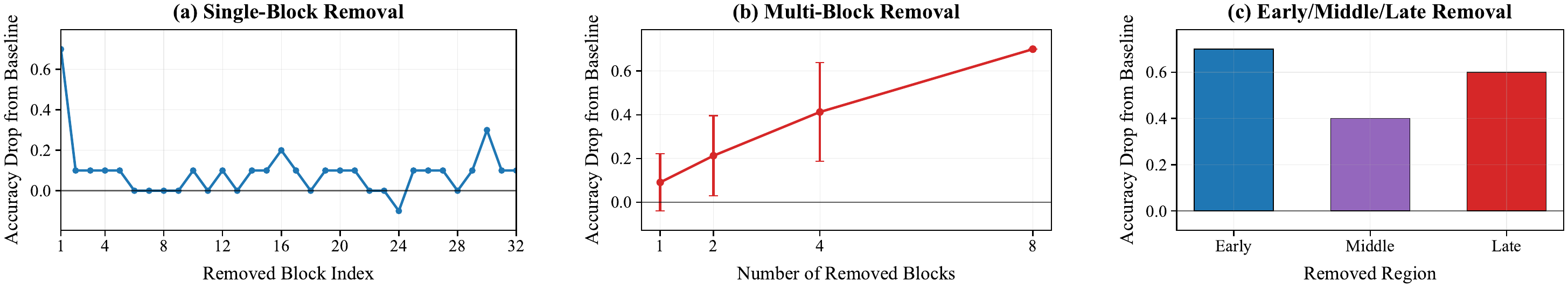}
    \caption{Block-removal perturbations. The diagnostic run has baseline accuracy $70\%$ on 10 GSM8K examples. Single-block and short-window removals often degrade gradually, while larger structured removals can collapse.}
    \label{fig:obs3}
\end{figure}

\begin{wrapfigure}{R}{0.5\textwidth}
    \centering
    \includegraphics[width=\linewidth]{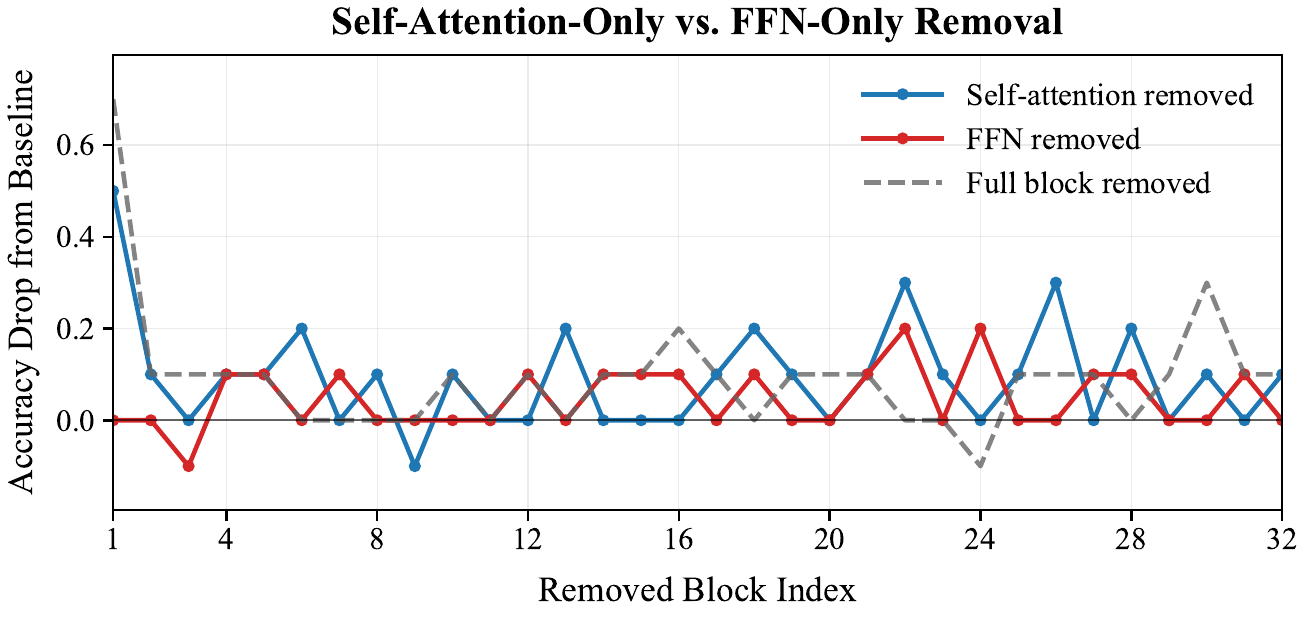}
    \caption{Self-attention-only and FFN-only removal diagnostics.}
    \label{fig:obs3-self-attention-ffn}
\end{wrapfigure}

\paragraph{Single-block removal.} Figure~\ref{fig:obs3}(a) removes one transformer block at a time. Most removals preserve nonzero performance, with mean accuracy $60.9\%$ versus the $70.0\%$ baseline. Thus no single middle block alone carries the whole denoising computation, although some block positions are more sensitive than others.

\paragraph{Consecutive multi-block removal.} Figure~\ref{fig:obs3}(b) removes consecutive windows of $1$, $2$, $4$, or $8$ blocks. Removing two or four blocks lowers mean accuracy to $48.8\%$ and $28.8\%$, while removing eight collapses the 10-example run. This does not mean layers are redundant. Rather, refinement is distributed across blocks, but still depends on preserving enough of the ordered layer trajectory.

\paragraph{Early, middle, and late region removal.} Figure~\ref{fig:obs3}(c) removes coarse four-block regions. Removing blocks 1--4 collapses accuracy to $0\%$, removing blocks 15--18 leaves $30\%$, and removing blocks 29--32 leaves $10\%$. Thus early representation formation and late decoding-oriented refinement are both important, while different regions affect the computation in different ways.

\paragraph{Self-attention-only vs. FFN-only removal.} Figure~\ref{fig:obs3-self-attention-ffn} removes only the self-attention or FFN sublayer at each block. The pattern again supports distributed refinement across many blocks and both sublayers, rather than a single isolated component carrying all denoising computation. The main exception is the first self-attention sublayer, whose removal causes a disproportionately large degradation, suggesting that most blocks behave like incremental refinement steps while the first self-attention sublayer forms the initial context-mixed state.

\subsection{Previous-Step Final States Are Consistent}
\label{sec:obs-cross-step}

The previous observations support layer-wise refinement within one denoising step. We next measure whether the final hidden state from one step remains aligned with the next step's final hidden state:
\begin{equation}
    \mathrm{CrossSim}_{s}
    =
    \mathbb{E}_{i}
    \left[
    \cos\left(H_{s-1,i}^{L}, H_{s,i}^{L}\right)
    \right].
\end{equation}

\begin{wrapfigure}{R}{0.5\textwidth}
    \centering
    \includegraphics[width=\linewidth]{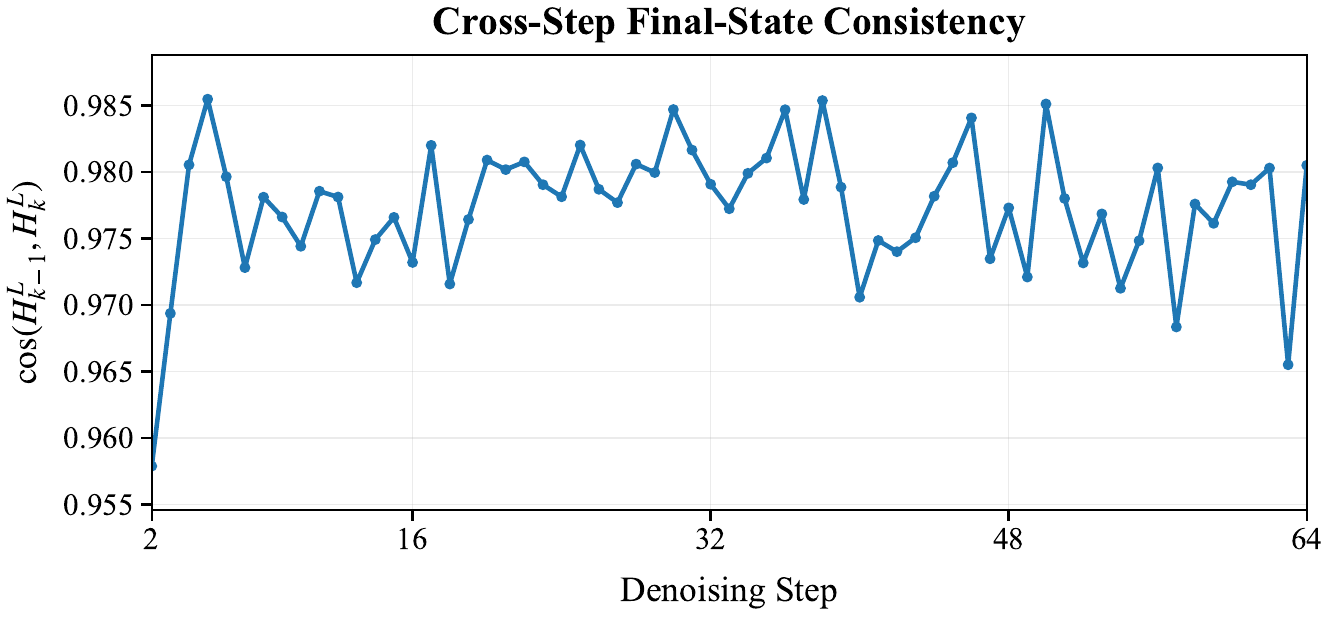}
    \caption{Final hidden states remain highly consistent across consecutive denoising steps.}
    \label{fig:obs4}
\end{wrapfigure}

\paragraph{Cross-step final-state consistency.} Figure~\ref{fig:obs4} measures cosine similarity between final hidden states at steps $s-1$ and $s$. Across steps 2--64, the mean valid-token similarity remains roughly $0.96$--$0.99$. The previous final hidden state is therefore not an unrelated activation, but already close to the next step's refinement state and a plausible source of cross-step continuous context.

\paragraph{Takeaway.} Together, the diagnostics suggest that MDLMs refine text across denoising steps in token space and across transformer layers in continuous hidden-state space. Standard sampling carries token updates forward but discards the continuous refinement trajectory, motivating a cross-step path for hidden-state refinement.

\section{Cross-Step Recurrent Refinement}
\label{sec:method}

The observations above suggest that MDLM generation contains both discrete token refinement across denoising steps and continuous hidden-state refinement inside each model call. Standard sampling preserves only the former, since each step receives the updated token sequence but discards the hidden-state trajectory computed in the previous step. Cross-Step Recurrent Refinement (CSRR) adds a lightweight recurrent path for this missing continuous state while leaving the original masked-token update rule unchanged.

\subsection{Formulation}

At denoising step $s$, a standard MDLM maps the current partially masked sequence to a layer-wise hidden-state trajectory and then updates tokens according to the sampler:
\begin{equation}
    x_s \rightarrow H_s^0,\dots,H_s^L
    \rightarrow p_\theta(x_0 \mid x_s)
    \rightarrow x_{s+1}.
\end{equation}
After the token update, $H_s^L$ is normally discarded. CSRR instead reuses the detached final-block hidden state from the previous denoising step.
\begin{equation}
    M_s = \mathrm{SG}\left(H_{s-1}^{L}\right),
\end{equation}
where $\mathrm{SG}$ denotes stop-gradient. At the first denoising step, no previous hidden state is available, so the model runs exactly as the original MDLM.

CSRR incorporates the previous-step hidden state into the first transformer block of the current denoising step. The insertion point is placed after the ordinary self-attention sublayer and before the FFN sublayer. Let $A_s^1$ denote the intermediate hidden state after the self-attention update in the first block. CSRR attends from this current-step state to the cached previous-step hidden state.
\begin{align}
    Q_s^1 &= \mathcal{N}^{\mathrm{attn}}_1(A_s^1) W_Q^{\mathrm{cross}}, \\
    K_s &= \mathcal{N}^{\mathrm{attn}}_1(M_s) W_K^{\mathrm{cross}}, \\
    V_s &= \mathcal{N}^{\mathrm{attn}}_1(M_s) W_V^{\mathrm{cross}}, \\
    C_s^1 &= \mathrm{Attn}(Q_s^1, K_s, V_s) W_O^{\mathrm{cross}}.
\end{align}
The resulting cross-attention output is added as a scaled residual.
\begin{equation}
    \widetilde{A}_s^1 = A_s^1 + \alpha C_s^1.
\end{equation}
The first block then feeds $\widetilde{A}_s^1$ into its ordinary FFN sublayer.
The full CSRR inference procedure is summarized in Appendix~\ref{sec:csrr-inference-algorithm}.

Training follows the original supervised denoising objective while exposing the model to consecutive denoising forward passes. After each forward pass, the detached final-block hidden state is reused by the next forward pass, matching the recurrent computation used at inference time. For a training segment with $K$ supervised denoising forward passes, let $p_\theta^{(k)}$ denote the predicted distribution from the $k$-th forward pass over supervised token positions, and let $x_{\mathrm{gold}}$ denote the corresponding clean targets. We minimize the averaged cross entropy:
\begin{equation}
    \mathcal{L}_{\mathrm{CSRR}}
    =
    \frac{1}{K}
    \sum_{k=1}^{K}
    \mathrm{CE}\left(p_\theta^{(k)}, x_{\mathrm{gold}}\right).
\end{equation}

\begin{table}[htbp]
    \caption{Main reasoning results on LLaDA-8B-Base and LLaDA-8B-Instruct. Unless otherwise specified, CSRR denotes the joint-training setting. The module-only variant trains only the newly introduced modules. All values are percentages. Columns under each benchmark denote generation length. Parenthesized values denote absolute improvements over the corresponding baseline.}
    \label{tab:main-reasoning-results}
    \centering
    \small
    \setlength{\tabcolsep}{4pt}
    \resizebox{\textwidth}{!}{
    \begin{tabular}{llcccccc}
        \toprule
        \multirow{2}{*}{NFE/Gen Len} & \multirow{2}{*}{Method}
        & \multicolumn{2}{c}{Countdown}
        & \multicolumn{2}{c}{GSM8K}
        & \multicolumn{2}{c}{SVAMP} \\
        \cmidrule(lr){3-4}\cmidrule(lr){5-6}\cmidrule(lr){7-8}
        & & 128 & 256 & 128 & 256 & 128 & 256 \\
        \midrule
        \rowcolor{gray!12}
        \multicolumn{8}{l}{\textit{LLaDA-8B-Base}} \\
        \multirow{3}{*}{$1/2$} & Baseline & 7.42 & 10.16 & 17.06 & 60.88 & 48.00 & 77.67 \\
        & CSRR (module only) & 15.23 {\scriptsize(\textcolor{green!50!black}{+7.81})} & 13.67 {\scriptsize(\textcolor{green!50!black}{+3.51})} & 19.03 {\scriptsize(\textcolor{green!50!black}{+1.97})} & 63.38 {\scriptsize(\textcolor{green!50!black}{+2.50})} & 56.33 {\scriptsize(\textcolor{green!50!black}{+8.33})} & 76.67 {\scriptsize(\textcolor{gray!55}{-1.00})} \\
        & \textbf{CSRR} & 17.58 {\scriptsize(\textcolor{green!50!black}{+10.16})} & 17.58 {\scriptsize(\textcolor{green!50!black}{+7.42})} & 33.59 {\scriptsize(\textcolor{green!50!black}{+16.53})} & 70.66 {\scriptsize(\textcolor{green!50!black}{+9.78})} & 66.67 {\scriptsize(\textcolor{green!50!black}{+18.67})} & 81.00 {\scriptsize(\textcolor{green!50!black}{+3.33})} \\
        \midrule
        \multirow{3}{*}{$1$} & Baseline & 6.25 & 11.72 & 16.68 & 64.75 & 52.00 & 80.33 \\
        & CSRR (module only) & 16.02 {\scriptsize(\textcolor{green!50!black}{+9.77})} & 12.50 {\scriptsize(\textcolor{green!50!black}{+0.78})} & 27.07 {\scriptsize(\textcolor{green!50!black}{+10.39})} & 70.13 {\scriptsize(\textcolor{green!50!black}{+5.38})} & 54.33 {\scriptsize(\textcolor{green!50!black}{+2.33})} & 78.33 {\scriptsize(\textcolor{gray!55}{-2.00})} \\
        & \textbf{CSRR} & 18.36 {\scriptsize(\textcolor{green!50!black}{+12.11})} & 21.88 {\scriptsize(\textcolor{green!50!black}{+10.16})} & 37.38 {\scriptsize(\textcolor{green!50!black}{+20.70})} & 79.23 {\scriptsize(\textcolor{green!50!black}{+14.48})} & 71.33 {\scriptsize(\textcolor{green!50!black}{+19.33})} & 82.67 {\scriptsize(\textcolor{green!50!black}{+2.34})} \\
        \midrule
        \rowcolor{gray!12}
        \multicolumn{8}{l}{\textit{LLaDA-8B-Instruct}} \\
        \multirow{3}{*}{$1/2$} & Baseline & 17.19 & 13.67 & 68.31 & 76.04 & 83.67 & 85.67 \\
        & CSRR (module only) & 23.83 {\scriptsize(\textcolor{green!50!black}{+6.64})} & 20.31 {\scriptsize(\textcolor{green!50!black}{+6.64})} & 68.92 {\scriptsize(\textcolor{green!50!black}{+0.61})} & 76.35 {\scriptsize(\textcolor{green!50!black}{+0.31})} & 86.33 {\scriptsize(\textcolor{green!50!black}{+2.66})} & 85.67 {\scriptsize(\textcolor{gray!55}{+0.00})} \\
        & \textbf{CSRR} & 28.52 {\scriptsize(\textcolor{green!50!black}{+11.33})} & 24.22 {\scriptsize(\textcolor{green!50!black}{+10.55})} & 70.58 {\scriptsize(\textcolor{green!50!black}{+2.27})} & 78.54 {\scriptsize(\textcolor{green!50!black}{+2.50})} & 86.00 {\scriptsize(\textcolor{green!50!black}{+2.33})} & 86.00 {\scriptsize(\textcolor{green!50!black}{+0.33})} \\
        \midrule
        \multirow{3}{*}{$1$} & Baseline & 23.44 & 21.48 & 70.43 & 78.62 & 87.00 & 86.67 \\
        & CSRR (module only) & 23.05 {\scriptsize(\textcolor{gray!55}{-0.39})} & 23.44 {\scriptsize(\textcolor{green!50!black}{+1.96})} & 71.95 {\scriptsize(\textcolor{green!50!black}{+1.52})} & 80.06 {\scriptsize(\textcolor{green!50!black}{+1.44})} & 89.67 {\scriptsize(\textcolor{green!50!black}{+2.67})} & 88.33 {\scriptsize(\textcolor{green!50!black}{+1.66})} \\
        & \textbf{CSRR} & 38.28 {\scriptsize(\textcolor{green!50!black}{+14.84})} & 29.30 {\scriptsize(\textcolor{green!50!black}{+7.82})} & 74.00 {\scriptsize(\textcolor{green!50!black}{+3.57})} & 83.47 {\scriptsize(\textcolor{green!50!black}{+4.85})} & 89.00 {\scriptsize(\textcolor{green!50!black}{+2.00})} & 89.00 {\scriptsize(\textcolor{green!50!black}{+2.33})} \\
        \bottomrule
    \end{tabular}
    }
\end{table}

\subsection{Design Rationale}

The implemented configuration uses the previous final-block hidden state and incorporates it after the first self-attention update of the current step. The previous final-block state is motivated by the cross-step consistency diagnostic, where consecutive final hidden states remain highly aligned. The insertion site is motivated by the block- and sublayer-removal results, where the first self-attention sublayer acts as a special context-forming component while later blocks largely follow the incremental refinement pattern.
Appendix~\ref{sec:app-source-injection-positions} further sweeps the source layer used for extraction and the injection layer used for incorporation, showing that CSRR remains effective for many layer pairs and that the default choice is among the stronger settings.

\subsection{Empirical Validation}

\subsubsection{Experimental Setup}
\label{sec:experimental-setup}

We mainly evaluate CSRR on LLaDA-8B-Base and LLaDA-8B-Instruct~\citep{llada}, and further conduct extension tests on Dream-7B-Base and Dream-7B-Instruct~\citep{dream2025}. For each backbone, we compare against the same sampler without cross-step hidden-state reuse. CSRR is fine-tuned on the S1K dataset~\citep{dataset-s1k} using consecutive denoising forward passes. 
We mainly evaluate reasoning benchmarks including Countdown~\citep{dataset-countdown}, GSM8K~\citep{dataset-gsm8k}, and SVAMP~\citep{dataset-svamp}, and additionally use HumanEval~\citep{dataset-humaneval}, HumanEval+~\citep{dataset-humaneval-plus}, and MBPP~\citep{dataset-mbpp} as code evaluation benchmarks. 
More implementation details and the evaluation prompt templates are listed in Appendices~\ref{sec:more-implementation-details} and~\ref{sec:evaluation-prompts}, respectively.

\subsubsection{Evaluation Results}

\paragraph{Main reasoning results.} Table~\ref{tab:main-reasoning-results} reports the main results on Countdown, GSM8K, and SVAMP using LLaDA-8B-Base and LLaDA-8B-Instruct. Unless otherwise specified, CSRR refers to the joint-training setting that trains both the newly introduced modules and LoRA adapters on the original backbone. CSRR improves the baseline across all reported reasoning configurations, with particularly large gains on Countdown and GSM8K for the base model. We also include a module-only variant, which often improves over the baseline and indicates that the added recurrent path contributes useful information even when only the new modules are trained.

\finishfigurewrap
\noindent\begin{minipage}[t]{\dimexpr0.5\textwidth-\columnsep\relax}
\vspace{0pt}
\paragraph{Code benchmarks.} Table~\ref{tab:code-benchmark-results} evaluates CSRR on HumanEval, HumanEval+, and MBPP using LLaDA-8B-Instruct. CSRR consistently improves pass rates across all reported code-generation settings, showing that the proposed hidden-state reuse is not limited to mathematical reasoning tasks.
\end{minipage}\hfill
\begin{minipage}[t]{0.5\textwidth}
\vspace{0pt}
    \captionof{table}{Code-generation results on LLaDA-8B-Instruct. NFE equals generation length in all code settings. All values are percentages, and $\Delta$ denotes the absolute improvement over the baseline.}
    \label{tab:code-benchmark-results}
    \centering
    \small
    \setlength{\tabcolsep}{3pt}
    \resizebox{\linewidth}{!}{
    \begin{tabular}{lrrrrrr}
        \toprule
        \multirow{2}{*}{Code benchmark}
        & \multicolumn{3}{c}{Gen Len 128}
        & \multicolumn{3}{c}{Gen Len 256} \\
        \cmidrule(lr){2-4}\cmidrule(lr){5-7}
        & Baseline & CSRR & $\Delta$ & Baseline & CSRR & $\Delta$ \\
        \midrule
        HumanEval & 27.44 & 29.88 & \textcolor{green!50!black}{+2.44} & 37.20 & 43.90 & \textcolor{green!50!black}{+6.70} \\
        HumanEval+ & 20.12 & 22.56 & \textcolor{green!50!black}{+2.44} & 29.88 & 34.15 & \textcolor{green!50!black}{+4.27} \\
        MBPP & 49.03 & 50.58 & \textcolor{green!50!black}{+1.55} & 49.03 & 50.97 & \textcolor{green!50!black}{+1.94} \\
        \bottomrule
    \end{tabular}
    }
\end{minipage}\par\medskip

\begin{wrapfigure}{r}{0.5\textwidth}
    \centering
    \includegraphics[width=\linewidth]{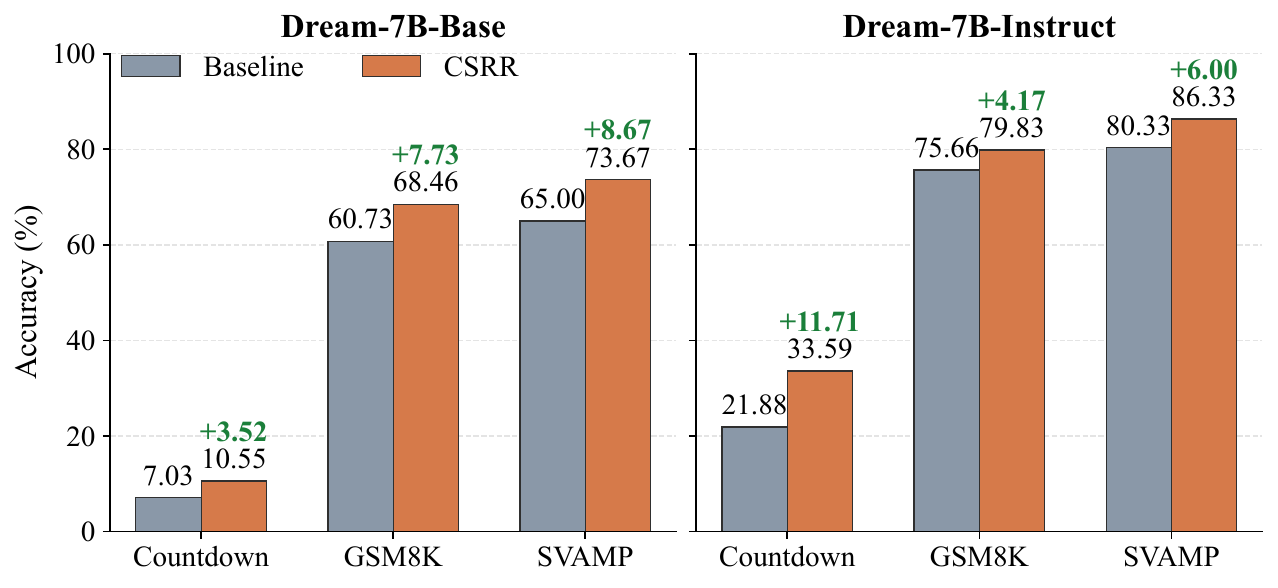}
    \caption{Reasoning results on Dream-7B-Base and Dream-7B-Instruct. All settings use generation length $256$ and NFE $128$.}
    \label{fig:dream-results}
\end{wrapfigure}

\paragraph{Extension to Dream models.} Figure~\ref{fig:dream-results} reports extension results on Dream-7B-Base and Dream-7B-Instruct. CSRR improves the corresponding baseline on all reported reasoning benchmarks, suggesting that the recurrent hidden-state path transfers beyond the LLaDA model family.

\finishfigurewrap
\noindent\begin{minipage}[t]{\dimexpr0.5\textwidth-\columnsep\relax}
\vspace{0pt}
\paragraph{KV-cache support.} Table~\ref{tab:kvcache-results} evaluates CSRR with KV-cache support on Countdown. The gains remain consistent across model variants and generation budgets, indicating that the recurrent hidden-state path is compatible with the cached inference setting. Full KV-cache results across Countdown, GSM8K, and SVAMP are provided in Appendix~\ref{sec:app-kvcache-results}.
\end{minipage}\hfill
\begin{minipage}[t]{0.5\textwidth}
\vspace{0pt}
    \captionof{table}{Countdown results with KV-cache support on LLaDA-8B-Base and LLaDA-8B-Instruct. All values are percentages, and $\Delta$ denotes the absolute improvement over the baseline.}
    \label{tab:kvcache-results}
    \centering
    \small
    \setlength{\tabcolsep}{4pt}
    \resizebox{\linewidth}{!}{
    \begin{tabular}{lrrrrrr}
        \toprule
        \multirow{2}{*}{NFE/Gen Len}
        & \multicolumn{3}{c}{Gen Len 128}
        & \multicolumn{3}{c}{Gen Len 256} \\
        \cmidrule(lr){2-4}\cmidrule(lr){5-7}
        & Baseline & CSRR & $\Delta$ & Baseline & CSRR & $\Delta$ \\
        \midrule
        \rowcolor{gray!12}
        \multicolumn{7}{l}{\textit{LLaDA-8B-Base}} \\
        $1/2$ & 8.59 & 17.19 & \textcolor{green!50!black}{+8.60} & 7.81 & 19.53 & \textcolor{green!50!black}{+11.72} \\
        $1$ & 7.03 & 22.66 & \textcolor{green!50!black}{+15.63} & 10.94 & 21.48 & \textcolor{green!50!black}{+10.54} \\
        \midrule
        \rowcolor{gray!12}
        \multicolumn{7}{l}{\textit{LLaDA-8B-Instruct}} \\
        $1/2$ & 17.19 & 25.39 & \textcolor{green!50!black}{+8.20} & 15.62 & 21.09 & \textcolor{green!50!black}{+5.47} \\
        $1$ & 21.48 & 26.56 & \textcolor{green!50!black}{+5.08} & 16.41 & 23.44 & \textcolor{green!50!black}{+7.03} \\
        \bottomrule
    \end{tabular}
    }
\end{minipage}\par\medskip

\begin{wrapfigure}{r}{0.5\textwidth}
    \centering
    \includegraphics[width=\linewidth]{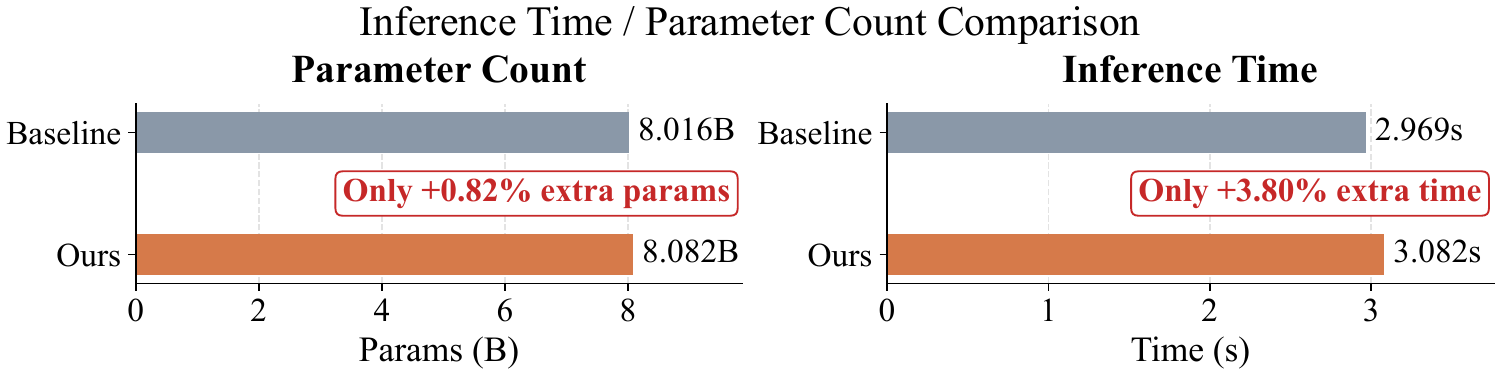}
    \caption{Inference-time and parameter-count comparison between the original LLaDA-8B-Instruct baseline and CSRR. Runtime is measured with the same prompt and sampling hyperparameters, using generation length 256 and NFE 128.}
    \label{fig:inference-param-comparison}
\end{wrapfigure}

\paragraph{Parameter and inference overhead.} Figure~\ref{fig:inference-param-comparison} compares the computational cost of CSRR against the original LLaDA-8B-Instruct sampler. The added cross-attention module increases the parameter count from $8.016$B to $8.082$B, corresponding to only $0.82\%$ additional parameters. With the same prompt and sampling hyperparameters, average inference time increases from $2.969$s to $3.082$s, a $3.80\%$ overhead. Thus, CSRR adds a small parameter footprint and modest runtime cost.

Additional S1K-matched fine-tuning controls, fixed-scale results, block-size results, and source-injection layer sweeps are provided in Appendices~\ref{sec:app-s1k-ft-control}, \ref{sec:app-fixed-scale-results}, \ref{sec:app-different-blocksize-results}, and~\ref{sec:app-source-injection-positions}, respectively.

\section{Related Work}

\paragraph{Masked diffusion language models.}
Diffusion language modeling~\citep{dLLM-survey,wu2025dmark,zhong2026efficient} adapts denoising diffusion~\citep{DDPM,song2019generative,podell2023sdxl,zhong2025outdreamer} to discrete token sequences. Early text generation models explored categorical corruption processes~\citep{d3pm}, BERT-style initialization and masked reconstruction~\citep{diffusionbert}, reparameterized discrete objectives~\citep{rdm}, and semi-autoregressive simplex diffusion~\citep{han2023ssd}. Masked diffusion language models (MDLMs)~\citep{mdlm,md4} further simplify this paradigm by formulating generation as iterative masked-token denoising, allowing each update to condition on bidirectional context rather than a strict left-to-right prefix. Recent DLMs have been scaled through instruction tuning~\citep{diffusion-llms}, adaptation from autoregressive backbones~\citep{DiffuGPT-DiffuLLaMA}, open model releases~\citep{tess-2,llada,dream2025,llada1.5,llada2}, and extensions to multimodal, reasoning, and long-context settings~\citep{yang2025mmada,zhao2025d1,liu2025longllada}. These works establish MDLMs as a practical alternative generation paradigm, but they mainly characterize denoising through token-level states or final output distributions.

\paragraph{Inference and latent refinement in DLMs.}
Another line of work improves the efficiency and flexibility of DLM sampling. Existing methods study blockwise or hybrid generation~\citep{arriolablock,sdar}, variable-length and sequential denoising~\citep{DAEDAL,liu2025sequential}, faster-than-autoregressive or parallel decoding~\citep{d2f,chen2025dparallel}, and cache-based or system-level acceleration~\citep{wu2025fast,wu2025fast2,liu2025dllm,ma2025dkv}. Recent variants also enrich hard remasking with soft masking~\citep{hersche2025soft}, coupled discrete-continuous diffusion~\citep{zheng2025continuously}, progressive token evolution~\citep{zhong2026beyond}, hierarchical semantic prediction~\citep{zhou2025next}, and latent reasoning trajectories~\citep{zhou2025coevolutionary,kang2025ladir}. These approaches modify the sampler, token-transition process, or explicit latent variables. Our work is complementary: we inspect the hidden-state trajectory that is already produced inside each MDLM forward pass, and use it as a continuous cross-step refinement signal without changing the masked-token update rule.

\paragraph{Recurrent and feedback transformers.}
Transformer recurrence has been explored as a way to increase effective computation, extend memory, or improve iterative reasoning. Universal and depth-recurrent transformers reuse blocks across depth or test-time computation~\citep{dehghani2018universal,geiping2026scaling,oncescu2026recurrent}, while segment-level and external-memory models carry information across long contexts~\citep{dai2019transformer,rae2019compressive,wu2022memorizing,bulatov2022recurrent,hutchins2022block}. Other architectures introduce feedback from previous time steps~\citep{fan2020addressing} or recurrent-style sequence modeling in transformer-like networks~\citep{sun2023retentive,peng2023rwkv}. CSRR shares the motivation of reusing continuous computation, but differs in setting and mechanism: it reuses the previous denoising step's hidden states in an MDLM, rather than adding recurrence over autoregressive tokens, long-context segments, or repeated depth alone.

\section{Conclusion}

In this paper, we studied hidden-state refinement in masked diffusion language models. Our diagnostics show that hidden states evolve smoothly across layers, become more predictive with depth, depend on distributed block-wise computation, and stay aligned across denoising steps. These findings reveal a continuous refinement trajectory that complements discrete token refinement. Motivated by this view, we proposed Cross-Step Recurrent Refinement, which carries previous-step hidden states through a lightweight cross-attention path while preserving masked-token sampling. Extensive experiments across multiple models and benchmarks show that CSRR improves MDLM generation with modest overhead.

\FloatBarrier
\bibliography{main}

\clearpage
\appendix
\renewcommand\thesection{\Alph{section}}
\renewcommand\thefigure{S\arabic{figure}}
\renewcommand\thetable{S\arabic{table}}
\renewcommand\theequation{S\arabic{equation}}
\makeatletter
\@ifundefined{thealgorithm}{}{\renewcommand\thealgorithm{S\arabic{algorithm}}}
\makeatother

\renewcommand\theHfigure{S\arabic{figure}}
\renewcommand\theHtable{S\arabic{table}}
\renewcommand\theHequation{S\arabic{equation}}
\makeatletter
\@ifundefined{theHalgorithm}{}{\renewcommand\theHalgorithm{S\arabic{algorithm}}}
\makeatother

\setcounter{figure}{0}
\setcounter{table}{0}
\setcounter{equation}{0}
\makeatletter
\@ifundefined{c@algorithm}{}{\setcounter{algorithm}{0}}
\makeatother

\section{Appendix Overview}
\label{sec:appendix-overview}

This appendix is organized to provide methodological details, implementation details, diagnostic definitions, additional hidden-state analyses, supplementary experimental results, and evaluation prompts.

\begin{itemize}
    \item \textbf{Appendix~\ref{sec:more-methodological-details}: More Methodological Details}
    \begin{itemize}
        \item Appendix~\ref{sec:csrr-inference-algorithm}: CSRR Inference Algorithm
    \end{itemize}
    \item \textbf{Appendix~\ref{sec:more-implementation-details}: More Implementation Details}
    \begin{itemize}
        \item Appendix~\ref{sec:more-implementation-details-dataset}: Detailed Descriptions of Training Dataset and Evaluation Benchmarks
        \item Appendix~\ref{sec:more-implementation-details-training}: Training Configurations for CSRR
        \item Appendix~\ref{sec:more-implementation-details-inference-eval}: Inference and Evaluation Setup
    \end{itemize}
    \item \textbf{Appendix~\ref{sec:detailed-diagnostic-definitions}: Detailed Diagnostic Definitions}
    \begin{itemize}
        \item Appendix~\ref{sec:diagnostic-token-groups}: Token Groups and Aggregation
        \item Appendix~\ref{sec:diagnostic-similarity}: Hidden-State Similarity Diagnostics
        \item Appendix~\ref{sec:diagnostic-early-exit}: Early-Exit Prediction Diagnostics
        \item Appendix~\ref{sec:diagnostic-correction}: Correction and Break Diagnostics
    \end{itemize}
    \item \textbf{Appendix~\ref{sec:more-hidden-state-refinement-analyses}: More Hidden-State Refinement Analyses}
    \begin{itemize}
        \item Appendix~\ref{sec:app-all-step-layer-smoothness}: All-Step Layer Smoothness
        \item Appendix~\ref{sec:app-task-model-smoothness}: Adjacent-Layer Smoothness Across Tasks and LLaDA Variants
        \item Appendix~\ref{sec:app-cka-layer-similarity}: CKA-Based Layer Similarity
        \item Appendix~\ref{sec:app-topk-confidence}: Early-Exit Top-k, Probability, and Entropy Diagnostics
        \item Appendix~\ref{sec:app-correction-break-rates}: Correction and Break Rates
    \end{itemize}
    \item \textbf{Appendix~\ref{sec:more-experimental-results}: More Experimental Results}
    \begin{itemize}
        \item Appendix~\ref{sec:app-kvcache-results}: KV-Cache Results Across Benchmarks
        \item Appendix~\ref{sec:app-s1k-ft-control}: S1K-Matched Fine-Tuning Control
        \item Appendix~\ref{sec:app-fixed-scale-results}: Fixed-Scale Results
        \item Appendix~\ref{sec:app-different-blocksize-results}: Results with Different Block Sizes
        \item Appendix~\ref{sec:app-source-injection-positions}: Source and Injection Layer Positions
    \end{itemize}
    \item \textbf{Appendix~\ref{sec:evaluation-prompts}: Evaluation Prompts}
    \begin{itemize}
        \item Appendix~\ref{sec:chat-template-wrapping}: Chat-Template Wrapping
        \item Appendix~\ref{sec:reasoning-prompts}: Reasoning Prompts
        \item Appendix~\ref{sec:code-generation-prompts}: Code-Generation Prompts
    \end{itemize}
\end{itemize}

\section{More Methodological Details}
\label{sec:more-methodological-details}

This section provides additional methodological details for CSRR, focusing on the inference procedure that carries a previous-step hidden state across denoising steps.

\subsection{CSRR Inference Algorithm}
\label{sec:csrr-inference-algorithm}

Algorithm~\ref{alg:csrr-inference} summarizes CSRR inference. The discrete token update is exactly the same sampler used by the corresponding MDLM baseline. The only additional state is the detached final hidden state from the previous denoising step, which is reused by the next step through the cross-step recurrent path described in the main paper.

\begin{algorithm}[t]
\caption{CSRR inference}
\label{alg:csrr-inference}
\begin{algorithmic}[1]
\State \textbf{Input} initial masked sequence $x_1$, denoising steps $S$, model $f_\theta$, sampler update rule $\mathrm{Update}$, recurrent scale $\alpha$
\State $M_1 \gets \varnothing$
\For{$s = 1,\ldots,S$}
    \If{$M_s = \varnothing$}
        \State $(Z_s, H_s^L) \gets f_\theta(x_s)$
    \Else
        \State $(Z_s, H_s^L) \gets f_\theta(x_s; M_s, \alpha)$
    \EndIf
    \State $x_{s+1} \gets \mathrm{Update}(x_s, Z_s)$
    \State $M_{s+1} \gets \mathrm{SG}(H_s^L)$
\EndFor
\State \Return $x_{S+1}$
\end{algorithmic}
\end{algorithm}

When $M_s$ is available, $f_\theta(x_s; M_s,\alpha)$ denotes the forward pass that injects $M_s$ after the first self-attention sublayer and before the first FFN sublayer. When $M_s$ is unavailable at the first step, the model reduces to the original MDLM forward pass. The sampler update rule may correspond to the same semi-autoregressive block sampling procedure used in the baseline, including low-confidence remasking and the selected NFE setting.

\section{More Implementation Details}
\label{sec:more-implementation-details}

This section provides implementation details for the experiments, including the training dataset, evaluation benchmarks, training configuration, and inference-time evaluation setup.

\subsection{Detailed Descriptions of Training Dataset and Evaluation Benchmarks}
\label{sec:more-implementation-details-dataset}

\noindent \textbf{Training dataset.}
We use a compact supervised dataset to adapt the recurrent hidden-state path while preserving the model's ability to produce explicit reasoning traces.
\begin{itemize}
    \item \textbf{S1K}~\citep{dataset-s1k}. We fine-tune CSRR on the S1K dataset, a compact reasoning dataset containing 1,000 diverse and difficult questions paired with distilled reasoning traces and solutions. We preserve the long-form reasoning format, which matches our goal of adapting the model to multi-step denoising trajectories rather than training it only on short final answers. The targets are formatted with an explicit reasoning field and a boxed final answer so that the model receives both intermediate solution structure and a clear answer boundary.
\end{itemize}

\noindent \textbf{Reasoning benchmarks.}
We evaluate mathematical reasoning with benchmarks that require different forms of multi-step inference, including symbolic arithmetic search, natural-language arithmetic, and robustness to problem rewrites.
\begin{itemize}
    \item \textbf{Countdown}~\citep{dataset-countdown}. Countdown evaluates arithmetic planning under explicit compositional constraints. Each instance provides a small set of numbers and a target value, and the model must construct an expression using the provided numbers and basic arithmetic operations to reach the target. This benchmark is useful for testing whether cross-step refinement improves search-like symbolic reasoning, since a valid answer must satisfy both the numerical target and the input-number constraints.
    \item \textbf{GSM8K}~\citep{dataset-gsm8k}. GSM8K is a grade-school math word-problem benchmark with linguistically diverse questions that typically require multi-step arithmetic reasoning. We evaluate on the test split and extract the final numeric answer from the generated solution. GSM8K serves as the main natural-language arithmetic benchmark because it combines textual understanding, intermediate calculation, and final-answer consistency.
    \item \textbf{SVAMP}~\citep{dataset-svamp}. SVAMP is a math word-problem benchmark designed to stress robustness to variations in problem wording and structure. Each example contains a short narrative, a question, an equation, and a numeric answer. Compared with GSM8K, SVAMP places more emphasis on avoiding shallow pattern matching, making it a complementary test of whether the model can maintain the intended semantic relations during iterative generation.
\end{itemize}

\noindent \textbf{Code benchmarks.}
We further evaluate code generation to test whether the method transfers from mathematical reasoning to executable program synthesis.
\begin{itemize}
    \item \textbf{HumanEval}~\citep{dataset-humaneval}. HumanEval evaluates Python function synthesis from a function signature and natural-language docstring. The model is asked to complete the function, and correctness is measured by executing the generated code against the provided unit tests. We use it to test whether the proposed hidden-state reuse mechanism transfers beyond mathematical reasoning to functional code generation.
    \item \textbf{HumanEval+}~\citep{dataset-humaneval-plus}. HumanEval+ follows the HumanEval problem format but evaluates generated programs with a substantially expanded test suite. It is a stricter functional-correctness benchmark because additional tests cover more edge cases and reduce the chance that a solution passes by matching only the original sparse tests.
    \item \textbf{MBPP}~\citep{dataset-mbpp}. MBPP contains short Python programming tasks described in natural language, with reference solutions and executable tests. We use the sanitized split, whose tasks are manually checked for clearer specifications and more reliable tests. MBPP complements HumanEval by emphasizing beginner-to-intermediate programming patterns such as list processing, string manipulation, arithmetic, and simple control flow.
\end{itemize}

\subsection{Training Configurations for CSRR}
\label{sec:more-implementation-details-training}

The main LLaDA backbones are LLaDA-8B-Base and LLaDA-8B-Instruct, and the extension settings additionally cover Dream-7B-Base and Dream-7B-Instruct. We consider two training regimes. In the main setting, we jointly train the newly introduced CSRR modules and lightweight LoRA adapters on the original backbone for 10,000 steps. The LoRA adapters are applied to the query, key, and value projections with rank $128$, alpha $256$, dropout $0.05$, and no bias tuning. We also report a module-only setting, where the pretrained backbone is frozen and only the newly added CSRR parameters are optimized for 20,000 steps. 
The main CSRR training run uses batch size $4$, maximum sequence length $1024$, learning rate $1\times10^{-5}$, weight decay $0.1$, and gradient clipping at $1.0$. The training collator simulates consecutive denoising forward passes with $K=2$. It samples a target block of length $512$, supervises tokens inside the selected block, updates one predicted token per forward stage, and samples the cross-attention scale from $\{0.5,1.0,1.5,2.0\}$ during training.

\subsection{Inference and Evaluation Setup}
\label{sec:more-implementation-details-inference-eval}

We evaluate CSRR with the same semi-autoregressive masked-token sampling procedure used by the corresponding baseline. Unless otherwise stated, generation uses block length $32$, low-confidence remasking, and generation temperature $0.5$. We evaluate cross-attention scales in $\{0.5,1.0,1.5,2.0\}$ and report the best value in the main results. 
We use NFE, short for number of function evaluations, to denote the total number of model forward evaluations during generation. We consider generation lengths $128$ and $256$, with NFE equal to either the generation length or half of it. 
For reasoning tasks, accuracy is computed after task-specific answer extraction. GSM8K and SVAMP use the extracted numeric answer, while Countdown additionally verifies that the generated expression reaches the target value using the provided numbers. For code tasks, we evaluate functional correctness by extracting the generated Python program and executing it against the benchmark tests. 

\section{Detailed Diagnostic Definitions}
\label{sec:detailed-diagnostic-definitions}

This section defines the main diagnostic quantities used in the main paper and the supplementary material. The definitions match the diagnostics reported in our analysis and use the final generated sequence as the pseudo clean target for early-exit prediction diagnostics.

\subsection{Token Groups and Aggregation}
\label{sec:diagnostic-token-groups}

Let $P$ denote the prompt length and let the current semi-autoregressive block span positions $[B_{\mathrm{start}}, B_{\mathrm{end}})$. We use \emph{prompt} for positions before $P$, \emph{generated} for generated positions before the current block end, \emph{valid} for all positions before the current block end, and \emph{current block} for positions inside $[B_{\mathrm{start}}, B_{\mathrm{end}})$. For early-exit diagnostics, the \emph{all} group refers to generated positions before the current block end, while \emph{masked} and \emph{unmasked} are generated positions that are respectively equal or not equal to the mask token before the current transfer step. We compute token-level quantities inside the selected group and aggregate them across recorded denoising steps and examples.

\subsection{Hidden-State Similarity Diagnostics}
\label{sec:diagnostic-similarity}

For a token group $G$, adjacent-layer cosine similarity is computed as
\begin{equation}
    \mathrm{Sim}_{s,\ell}(G)
    =
    \frac{1}{|G|}
    \sum_{i\in G}
    \cos\left(H_{s,i}^{\ell}, H_{s,i}^{\ell+1}\right).
\end{equation}
The layer-pair heatmap uses the same token-wise cosine computation for all layer pairs:
\begin{equation}
    \mathrm{PairSim}_{s,\ell,j}
    =
    \frac{1}{N}
    \sum_{i=1}^{N}
    \cos\left(H_{s,i}^{\ell}, H_{s,i}^{j}\right).
\end{equation}
The relative update magnitude normalizes the update size by the current hidden-state norm:
\begin{equation}
    r_{s,\ell}
    =
    \frac{\lVert H_s^{\ell+1} - H_s^\ell \rVert_2}
         {\lVert H_s^\ell \rVert_2}.
\end{equation}
For CKA, we form centered token-feature matrices $\bar{X}_s^\ell$ and $\bar{X}_s^j$ from selected token positions, compute Gram matrices $K_\ell=\bar{X}_s^\ell(\bar{X}_s^\ell)^\top$ and $K_j=\bar{X}_s^j(\bar{X}_s^j)^\top$, and report
\begin{equation}
    \mathrm{CKA}(\ell,j)
    =
    \frac{\langle K_\ell, K_j\rangle_F}
         {\lVert K_\ell\rVert_F \lVert K_j\rVert_F}.
\end{equation}

\subsection{Early-Exit Prediction Diagnostics}
\label{sec:diagnostic-early-exit}

For each intermediate hidden state $H_s^\ell$, we apply the model's final normalization and shared LM head to obtain intermediate logits:
\begin{equation}
    Z_s^\ell = W_{\mathrm{lm}}\mathcal{N}^{\mathrm{final}}(H_s^\ell).
\end{equation}
Let $y_i$ denote the token at position $i$ in the final generated sequence and let $p_{s,i}^\ell=\mathrm{Softmax}(Z_{s,i}^\ell)$. Early-exit accuracy and cross entropy over a token group $G$ are
\begin{align}
    \mathrm{Acc}_{s,\ell}(G)
    &=
    \frac{1}{|G|}
    \sum_{i\in G}
    \mathbf{1}\!\left[
    \arg\max_v p_{s,i}^\ell(v)=y_i
    \right], \\
    \mathrm{CE}_{s,\ell}(G)
    &=
    -\frac{1}{|G|}
    \sum_{i\in G}
    \log p_{s,i}^\ell(y_i).
\end{align}
The top-$k$ accuracy, target-token probability, maximum probability, and entropy are
\begin{small}
\begin{align}
    \mathrm{TopKAcc}_{s,\ell}^{(k)}(G)
    &=
    |G|^{-1}
    \sum_{i\in G}
    \mathbf{1}\!\left[
    y_i \in \mathrm{TopK}_k(p_{s,i}^\ell)
    \right], \\
    \mathrm{Prob}_{s,\ell}(G)
    &=
    |G|^{-1}
    \sum_{i\in G}
    p_{s,i}^\ell(y_i), \\
    \mathrm{MaxProb}_{s,\ell}(G)
    &=
    |G|^{-1}
    \sum_{i\in G}
    \max_v p_{s,i}^\ell(v), \\
    \mathrm{Ent}_{s,\ell}(G)
    &=
    -|G|^{-1}
    \sum_{i\in G}
    \sum_v
    p_{s,i}^\ell(v)\log p_{s,i}^\ell(v).
\end{align}
\end{small}
For update-direction alignment, we compare each local layer update with the remaining direction toward the final hidden state:
\begin{small}
\begin{align}
    \mathrm{Align}_{s,\ell}(G)
    &=
    |G|^{-1}
    \sum_{i\in G}
    \cos\left(\Delta H_{s,i}^{\ell}, R_{s,i}^{\ell}\right), \\
    \Delta H_{s,i}^{\ell}
    &=
    H_{s,i}^{\ell+1}-H_{s,i}^{\ell},
    \quad
    R_{s,i}^{\ell}=H_{s,i}^{L}-H_{s,i}^{\ell}.
\end{align}
\end{small}

\subsection{Correction and Break Diagnostics}
\label{sec:diagnostic-correction}

Let $c_{s,\ell,i}=\mathbf{1}[\arg\max_v p_{s,i}^\ell(v)=y_i]$ denote whether layer $\ell$ predicts the pseudo clean target at token $i$. The fix rate measures wrong-to-correct transitions between adjacent layers, while the break rate measures correct-to-wrong transitions:
\begin{small}
\begin{align}
    \mathrm{Fix}_{s,\ell}(G)
    &=
    |G|^{-1}
    \sum_{i\in G}
    \mathbf{1}\!\left[
    \neg c_{s,\ell,i}
    \wedge c_{s,\ell+1,i}
    \right], \\
    \mathrm{Break}_{s,\ell}(G)
    &=
    |G|^{-1}
    \sum_{i\in G}
    \mathbf{1}\!\left[
    c_{s,\ell,i}
    \wedge \neg c_{s,\ell+1,i}
    \right].
\end{align}
\end{small}
We also report the net correction rate $\mathrm{Fix}_{s,\ell}(G)-\mathrm{Break}_{s,\ell}(G)$, which indicates whether an adjacent layer pair repairs more predictions than it disrupts.

\begin{figure}[htbp]
    \centering
    \includegraphics[width=\textwidth]{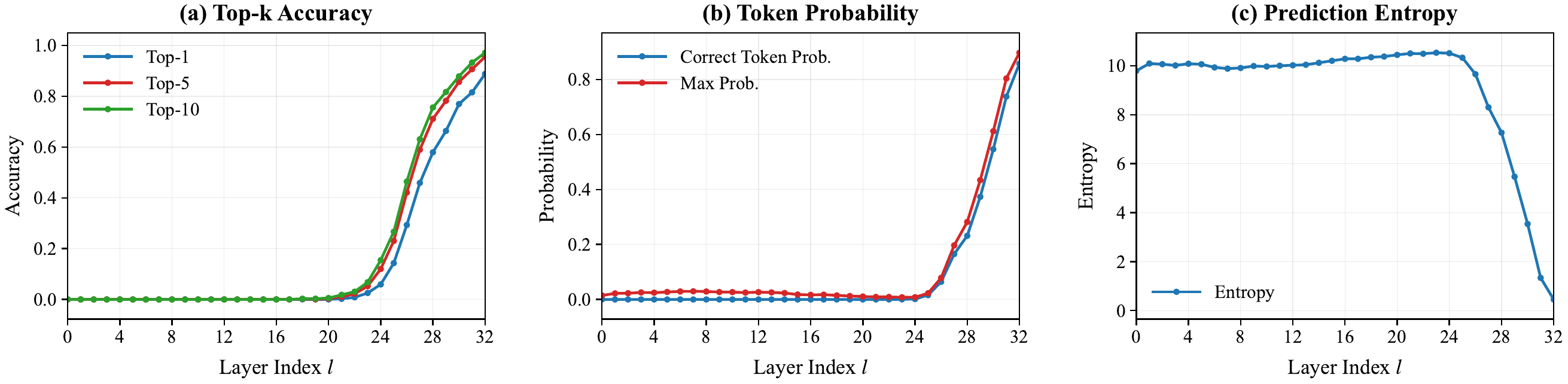}
    \caption{Top-$k$ accuracy, token probability, and entropy diagnostics for intermediate-layer predictions.}
    \label{fig:app-obs2-topk-entropy}
\end{figure}

\section{More Hidden-State Refinement Analyses}
\label{sec:more-hidden-state-refinement-analyses}

This section provides additional observation figures that complement the main paper. The main text focuses on a compact set of diagnostics so that the overall argument remains readable. Here, we include supporting analyses that examine the same hidden-state refinement view with broader denoising-step coverage, different tasks and model variants, an alternative representation-similarity metric, and finer-grained early-exit prediction statistics. These supplementary diagnostics broaden the empirical basis of the main observations while keeping the core discussion compact.

\Needspace{6\baselineskip}
\subsection{All-Step Layer Smoothness}
\label{sec:app-all-step-layer-smoothness}

\begin{wrapfigure}{r}{0.5\textwidth}
    \centering
    \includegraphics[width=\linewidth]{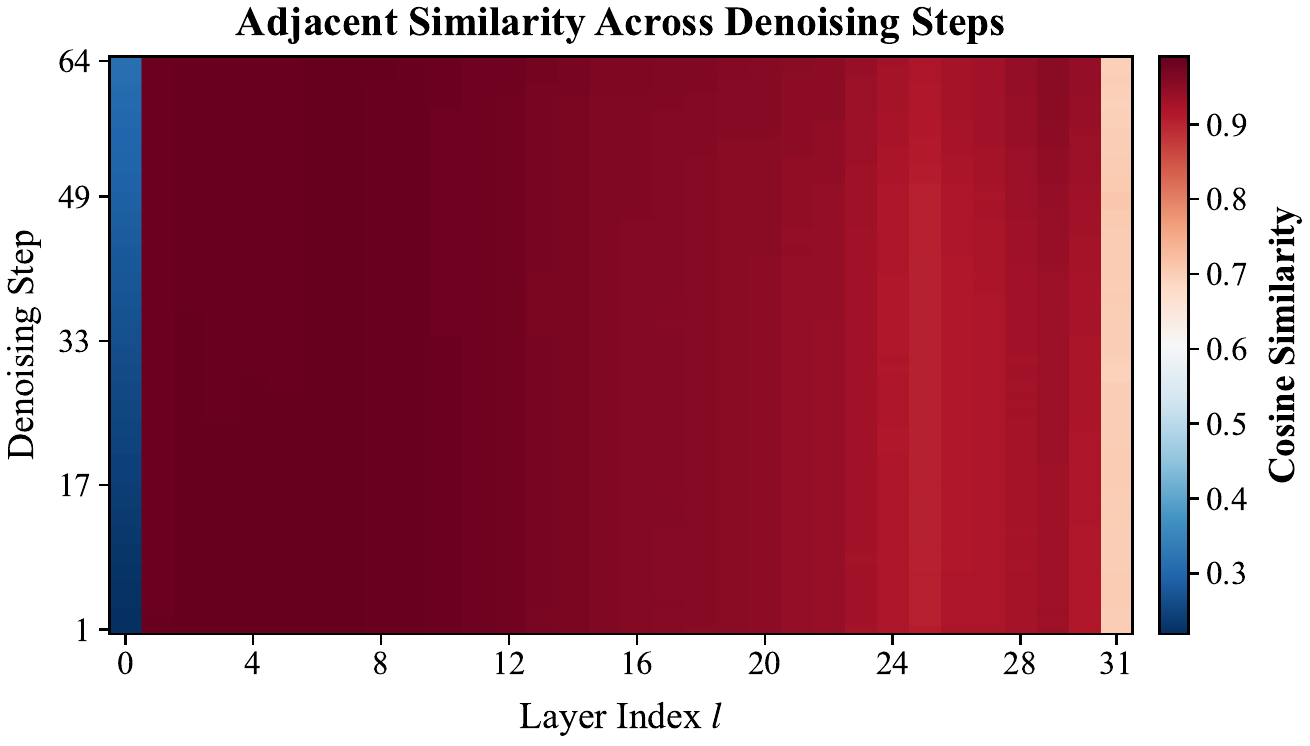}
    \caption{Adjacent-layer cosine similarity across all denoising steps.}
    \label{fig:app-obs1-all-step-heatmap}
\end{wrapfigure}

Figure~\ref{fig:app-obs1-all-step-heatmap} reports adjacent-layer cosine similarity across all denoising steps rather than only the selected steps shown in the main text. Each cell aggregates the token-wise cosine similarity between $H_s^\ell$ and $H_s^{\ell+1}$ at a particular denoising step and adjacent layer pair. This view is useful because selected-step curves can hide transient behavior. If the layer-wise refinement pattern only appeared at a few steps, the heatmap would show isolated high-similarity regions instead of a coherent structure over the full trajectory. In practice, the high-similarity band persists across most denoising steps, especially through the middle layers. The main deviation remains the transition from the embedding state into the first layer, which is consistent with the main-text interpretation that the first block has a stronger role in forming an initial contextual state before later layers perform more incremental updates.

\subsection{Adjacent-Layer Smoothness Across Tasks and LLaDA Variants}
\label{sec:app-task-model-smoothness}

\begin{wrapfigure}{r}{0.5\textwidth}
    \centering
    \includegraphics[width=\linewidth]{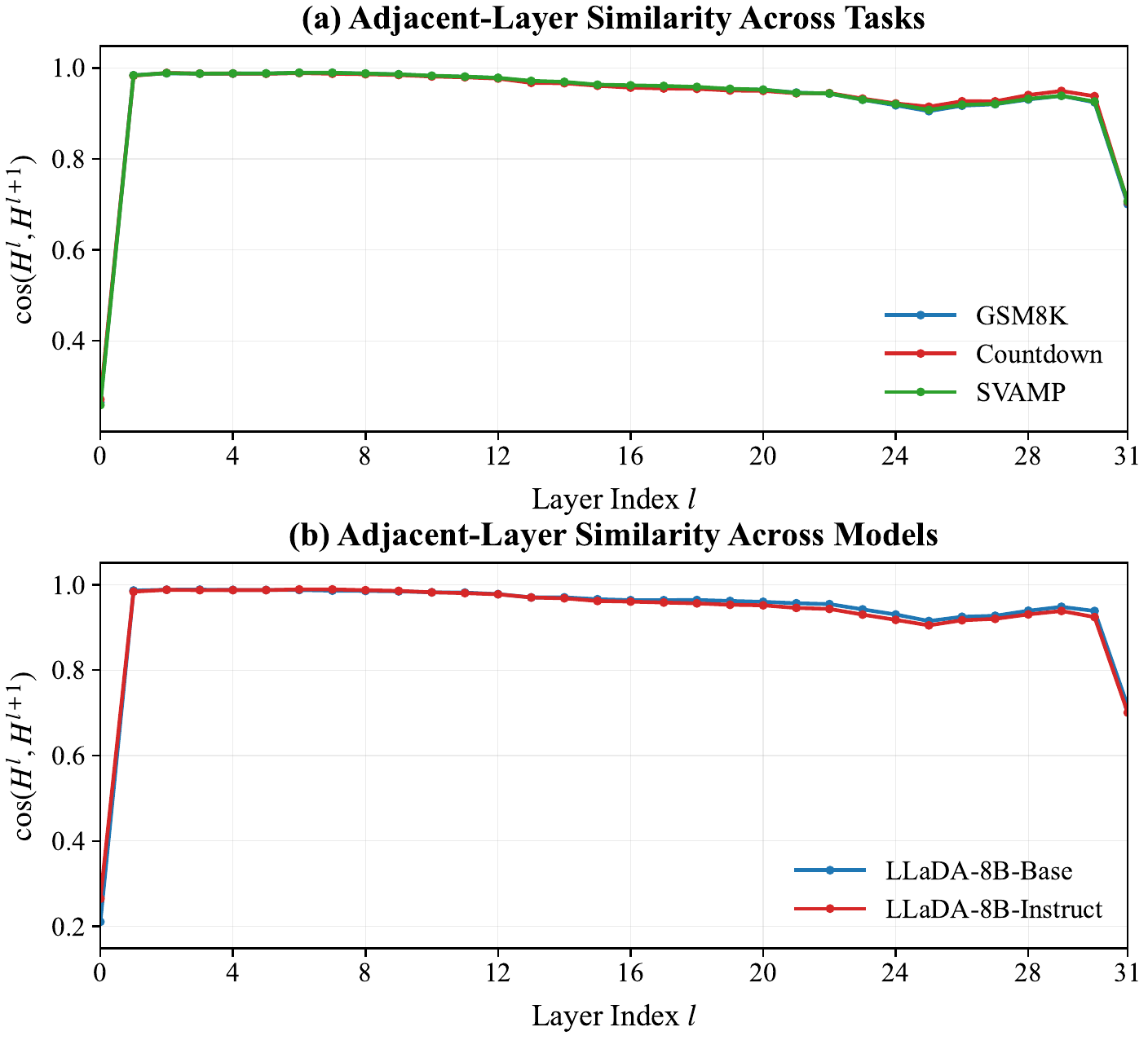}
    \caption{Adjacent-layer cosine similarity across tasks and model variants.}
    \label{fig:app-obs1-tasks-models}
\end{wrapfigure}

Figure~\ref{fig:app-obs1-tasks-models} extends the adjacent-layer similarity diagnostic beyond the primary setup used in the main observation, which uses LLaDA-8B-Instruct on GSM8K. We use these additional runs to examine whether the main smoothness pattern persists across different prompt distributions and model variants. The task panel compares GSM8K, Countdown, and SVAMP under the same observation protocol, covering natural-language arithmetic, arithmetic search, and robustness-oriented word problems. The model panel compares LLaDA-8B-Base with LLaDA-8B-Instruct, which isolates whether instruction tuning substantially changes the layer-wise continuity pattern. Across these settings, the curves remain high and have broadly similar shapes, suggesting that adjacent-layer smoothness is not a peculiarity of one benchmark split or one model variant.

\finishfigurewrap
\noindent\begin{minipage}{\linewidth}
\subsection{CKA-Based Layer Similarity}
\label{sec:app-cka-layer-similarity}
\noindent\begin{minipage}[t]{\dimexpr0.5\linewidth-\columnsep\relax}
\vspace{0pt}
Figure~\ref{fig:app-obs1-cka} provides a complementary representation-similarity view using linear CKA. Unlike token-wise cosine similarity, which compares each token representation directly, CKA compares the geometry of token-feature matrices after centering. This makes it a useful sanity check because it asks whether layers preserve similar relational structure among token representations, not only whether corresponding token vectors point in similar directions. The diagnostic computes CKA at selected denoising steps and averages the resulting layer-pair similarities. The resulting heatmap again has a strong diagonal structure, with nearby layers more similar than distant layers. This agrees with the cosine-based heatmap in the main paper, suggesting that the smooth layer-wise trajectory is visible under both direct vector similarity and a representation-level similarity measure.
\end{minipage}\hfill
\begin{minipage}[t]{0.5\linewidth}
\vspace{0pt}
    \centering
    \includegraphics[width=\linewidth]{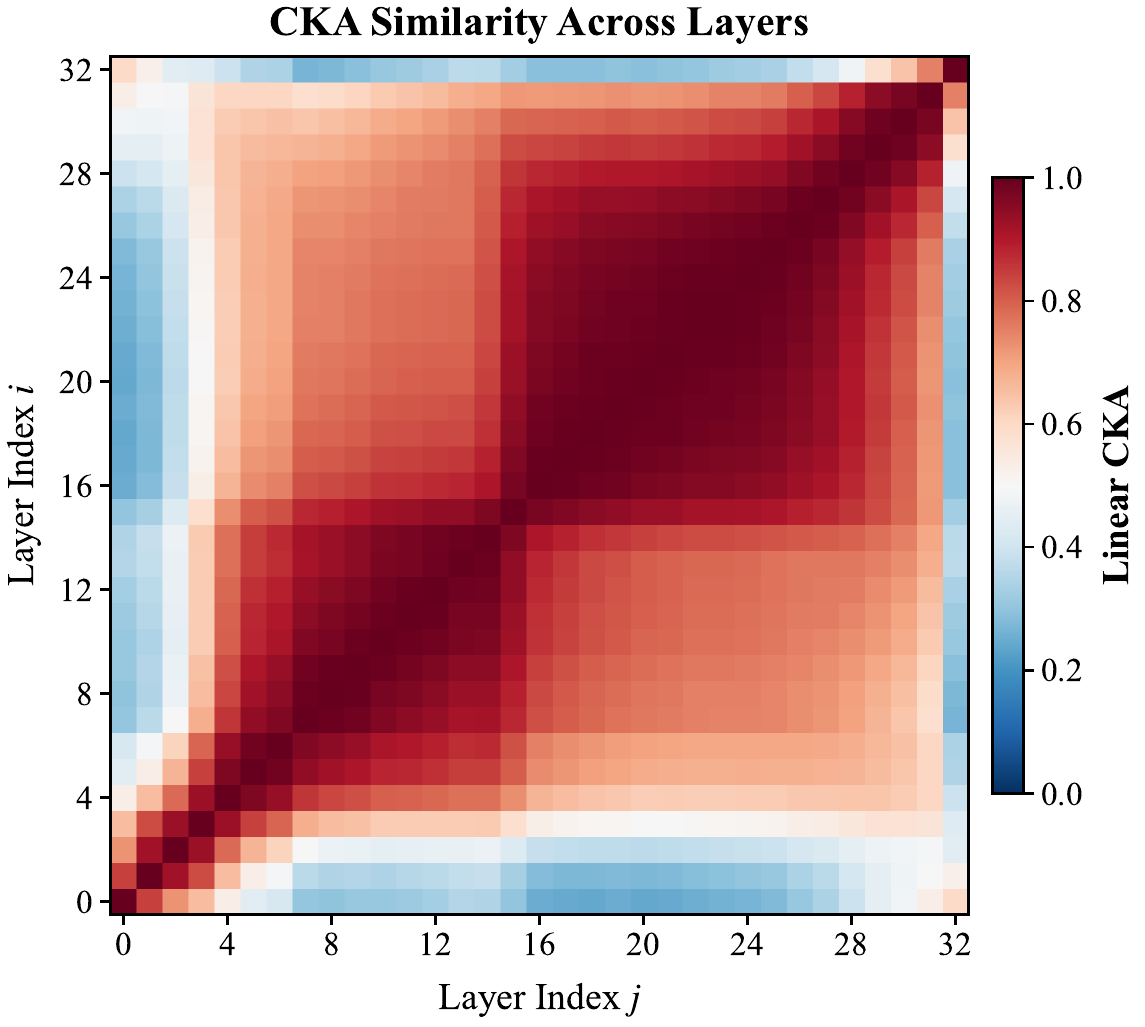}
    \captionof{figure}{Linear CKA similarity across transformer layers.}
    \label{fig:app-obs1-cka}
\end{minipage}
\end{minipage}\par

\subsection{Early-Exit Top-k, Probability, and Entropy Diagnostics}
\label{sec:app-topk-confidence}

Figure~\ref{fig:app-obs2-topk-entropy} adds distribution-level statistics to the early-exit analysis. For each intermediate layer, we apply the final normalization and shared LM head used by the model output layer, and compare predictions with the final generated sequence used as the pseudo clean target. The top-$k$ curves measure how early the target token becomes highly ranked, the probability curves track both target-token probability and maximum prediction probability, and the entropy curve measures the sharpness of the intermediate output distribution. These diagnostics give a more detailed picture than accuracy alone. Late layers do not merely flip a small number of argmax decisions. They also raise the rank and probability of the target token while making the distribution more decisive, supporting a gradual transformation from weak intermediate representations to output-ready denoising states, with the clearest change appearing in the final layers.

\begin{table}[htbp]
    \caption{Fixed-scale reasoning results with $\alpha=0.5$ on LLaDA-8B-Base and LLaDA-8B-Instruct. Columns under each benchmark denote generation length. All values are percentages. Parenthesized values denote absolute improvements over the corresponding baseline.}
    \label{tab:appendix-fixed-scale}
    \centering
    \small
    \setlength{\tabcolsep}{4pt}
    \resizebox{\textwidth}{!}{
    \begin{tabular}{llcccccc}
        \toprule
        \multirow{2}{*}{NFE/Gen Len} & \multirow{2}{*}{Method}
        & \multicolumn{2}{c}{Countdown}
        & \multicolumn{2}{c}{GSM8K}
        & \multicolumn{2}{c}{SVAMP} \\
        \cmidrule(lr){3-4}\cmidrule(lr){5-6}\cmidrule(lr){7-8}
        & & 128 & 256 & 128 & 256 & 128 & 256 \\
        \midrule
        \rowcolor{gray!12}
        \multicolumn{8}{l}{\textit{LLaDA-8B-Base}} \\
        \multirow{2}{*}{$1/2$} & Baseline & 7.42 & 10.16 & 17.06 & 60.88 & 48.00 & 77.67 \\
        & \textbf{CSRR} ($\alpha=0.5$) & 15.23 {\scriptsize(\textcolor{green!50!black}{+7.81})} & 17.58 {\scriptsize(\textcolor{green!50!black}{+7.42})} & 33.43 {\scriptsize(\textcolor{green!50!black}{+16.37})} & 70.66 {\scriptsize(\textcolor{green!50!black}{+9.78})} & 64.33 {\scriptsize(\textcolor{green!50!black}{+16.33})} & 81.00 {\scriptsize(\textcolor{green!50!black}{+3.33})} \\
        \midrule
        \multirow{2}{*}{$1$} & Baseline & 6.25 & 11.72 & 16.68 & 64.75 & 52.00 & 80.33 \\
        & \textbf{CSRR} ($\alpha=0.5$) & 18.36 {\scriptsize(\textcolor{green!50!black}{+12.11})} & 21.88 {\scriptsize(\textcolor{green!50!black}{+10.16})} & 37.38 {\scriptsize(\textcolor{green!50!black}{+20.70})} & 79.23 {\scriptsize(\textcolor{green!50!black}{+14.48})} & 70.00 {\scriptsize(\textcolor{green!50!black}{+18.00})} & 82.67 {\scriptsize(\textcolor{green!50!black}{+2.34})} \\
        \midrule
        \rowcolor{gray!12}
        \multicolumn{8}{l}{\textit{LLaDA-8B-Instruct}} \\
        \multirow{2}{*}{$1/2$} & Baseline & 17.19 & 13.67 & 68.31 & 76.04 & 83.67 & 85.67 \\
        & \textbf{CSRR} ($\alpha=0.5$) & 28.12 {\scriptsize(\textcolor{green!50!black}{+10.93})} & 20.70 {\scriptsize(\textcolor{green!50!black}{+7.03})} & 69.67 {\scriptsize(\textcolor{green!50!black}{+1.36})} & 78.54 {\scriptsize(\textcolor{green!50!black}{+2.50})} & 83.00 {\scriptsize(\textcolor{gray!55}{-0.67})} & 85.67 {\scriptsize(\textcolor{gray!55}{+0.00})} \\
        \midrule
        \multirow{2}{*}{$1$} & Baseline & 23.44 & 21.48 & 70.43 & 78.62 & 87.00 & 86.67 \\
        & \textbf{CSRR} ($\alpha=0.5$) & 38.28 {\scriptsize(\textcolor{green!50!black}{+14.84})} & 29.30 {\scriptsize(\textcolor{green!50!black}{+7.82})} & 74.00 {\scriptsize(\textcolor{green!50!black}{+3.57})} & 83.47 {\scriptsize(\textcolor{green!50!black}{+4.85})} & 89.00 {\scriptsize(\textcolor{green!50!black}{+2.00})} & 88.67 {\scriptsize(\textcolor{green!50!black}{+2.00})} \\
        \bottomrule
    \end{tabular}
    }
\end{table}

\Needspace{6\baselineskip}
\subsection{Correction and Break Rates}
\label{sec:app-correction-break-rates}

\begin{wrapfigure}{r}{0.5\textwidth}
    \centering
    \includegraphics[width=\linewidth]{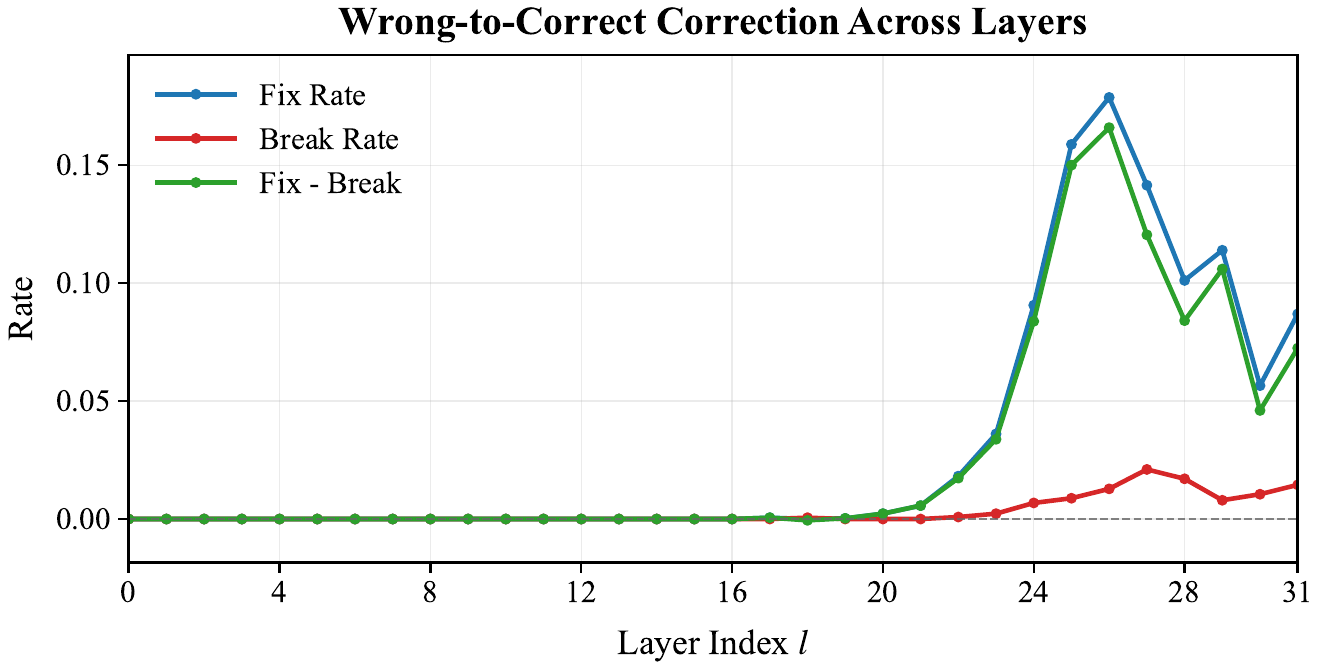}
    \caption{Wrong-to-correct fix rate, correct-to-wrong break rate, and their difference across adjacent layers.}
    \label{fig:app-obs2-correction-rate}
\end{wrapfigure}

Figure~\ref{fig:app-obs2-correction-rate} decomposes adjacent-layer prediction changes into fix and break events. A fix occurs when a token prediction is incorrect at layer $\ell$ but correct at layer $\ell+1$, while a break occurs when the reverse transition happens. This diagnostic is useful because an increase in final accuracy could in principle come from large unstable changes, where layers both fix many wrong tokens and break many correct tokens. The measured curves show a different pattern. In the later layers, the fix rate rises more strongly than the break rate, and the net correction curve becomes positive. This indicates that the layer-wise updates are not simply random reshuffling of token predictions. They are biased toward repairing intermediate mistakes while largely preserving predictions that have already become correct, which further supports the interpretation of the upper layers as refinement-oriented computation.

\section{More Experimental Results}
\label{sec:more-experimental-results}

This section provides additional experimental results that complement the main paper. We first report full KV-cache results across reasoning benchmarks, then compare CSRR with a standard fine-tuning control using the same S1K data and training budget. The main results select the best recurrent scale from the evaluated candidates, while the fixed-scale study below asks whether CSRR remains effective without per-setting scale selection. We also vary the semi-autoregressive block size to test whether the gains depend on a particular block granularity.

\subsection{KV-Cache Results Across Benchmarks}
\label{sec:app-kvcache-results}

Table~\ref{tab:appendix-kvcache-results} extends the KV-cache evaluation in the main paper from Countdown to all three reasoning benchmarks. CSRR consistently improves over the corresponding cached baseline across both LLaDA variants, both NFE settings, and all reported benchmarks. These results indicate that the recurrent hidden-state path remains compatible with cached inference beyond the Countdown setting shown in the main text.

\begin{table}[htbp]
    \caption{KV-cache reasoning results on LLaDA-8B-Base and LLaDA-8B-Instruct. Columns under each benchmark denote generation length. All values are percentages. Parenthesized values denote absolute improvements over the corresponding baseline.}
    \label{tab:appendix-kvcache-results}
    \centering
    \small
    \setlength{\tabcolsep}{4pt}
    \resizebox{\textwidth}{!}{
    \begin{tabular}{llcccccc}
        \toprule
        \multirow{2}{*}{NFE/Gen Len} & \multirow{2}{*}{Method}
        & \multicolumn{2}{c}{Countdown}
        & \multicolumn{2}{c}{GSM8K}
        & \multicolumn{2}{c}{SVAMP} \\
        \cmidrule(lr){3-4}\cmidrule(lr){5-6}\cmidrule(lr){7-8}
        & & 128 & 256 & 128 & 256 & 128 & 256 \\
        \midrule
        \rowcolor{gray!12}
        \multicolumn{8}{l}{\textit{LLaDA-8B-Base}} \\
        \multirow{2}{*}{$1/2$} & Baseline & 8.59 & 7.81 & 15.62 & 52.92 & 52.67 & 70.33 \\
        & \textbf{CSRR} & 17.19 {\scriptsize(\textcolor{green!50!black}{+8.60})} & 19.53 {\scriptsize(\textcolor{green!50!black}{+11.72})} & 28.35 {\scriptsize(\textcolor{green!50!black}{+12.73})} & 66.19 {\scriptsize(\textcolor{green!50!black}{+13.27})} & 60.67 {\scriptsize(\textcolor{green!50!black}{+8.00})} & 71.67 {\scriptsize(\textcolor{green!50!black}{+1.34})} \\
        \midrule
        \multirow{2}{*}{$1$} & Baseline & 7.03 & 10.94 & 16.45 & 62.85 & 53.33 & 78.67 \\
        & \textbf{CSRR} & 22.66 {\scriptsize(\textcolor{green!50!black}{+15.63})} & 21.48 {\scriptsize(\textcolor{green!50!black}{+10.54})} & 29.87 {\scriptsize(\textcolor{green!50!black}{+13.42})} & 74.98 {\scriptsize(\textcolor{green!50!black}{+12.13})} & 70.33 {\scriptsize(\textcolor{green!50!black}{+17.00})} & 81.67 {\scriptsize(\textcolor{green!50!black}{+3.00})} \\
        \midrule
        \rowcolor{gray!12}
        \multicolumn{8}{l}{\textit{LLaDA-8B-Instruct}} \\
        \multirow{2}{*}{$1/2$} & Baseline & 17.19 & 15.62 & 66.41 & 74.15 & 82.33 & 80.67 \\
        & \textbf{CSRR} & 25.39 {\scriptsize(\textcolor{green!50!black}{+8.20})} & 21.09 {\scriptsize(\textcolor{green!50!black}{+5.47})} & 67.17 {\scriptsize(\textcolor{green!50!black}{+0.76})} & 75.59 {\scriptsize(\textcolor{green!50!black}{+1.44})} & 86.67 {\scriptsize(\textcolor{green!50!black}{+4.34})} & 88.00 {\scriptsize(\textcolor{green!50!black}{+7.33})} \\
        \midrule
        \multirow{2}{*}{$1$} & Baseline & 21.48 & 16.41 & 69.14 & 78.01 & 85.33 & 84.67 \\
        & \textbf{CSRR} & 26.56 {\scriptsize(\textcolor{green!50!black}{+5.08})} & 23.44 {\scriptsize(\textcolor{green!50!black}{+7.03})} & 73.31 {\scriptsize(\textcolor{green!50!black}{+4.17})} & 81.65 {\scriptsize(\textcolor{green!50!black}{+3.64})} & 89.33 {\scriptsize(\textcolor{green!50!black}{+4.00})} & 90.33 {\scriptsize(\textcolor{green!50!black}{+5.66})} \\
        \bottomrule
    \end{tabular}
    }
\end{table}

\subsection{S1K-Matched Fine-Tuning Control}
\label{sec:app-s1k-ft-control}

\finishfigurewrap
\noindent\begin{minipage}[t]{\dimexpr0.5\textwidth-\columnsep\relax}
\vspace{0pt}
Table~\ref{tab:appendix-s1k-ft-control} compares CSRR with a standard fine-tuning control trained on the same S1K data for the same number of steps. The FT control improves over the untuned baseline, confirming that the training data itself is beneficial. CSRR remains stronger across all three benchmarks and both LLaDA variants, suggesting that the gains are not explained solely by matched supervised fine-tuning.
\end{minipage}\hfill
\begin{minipage}[t]{0.5\textwidth}
\vspace{0pt}
    \captionof{table}{S1K-matched fine-tuning control. Accuracy is reported in percentages with generation length $=$ NFE $=256$. FT denotes standard fine-tuning on the same S1K training set for the same number of steps.}
    \label{tab:appendix-s1k-ft-control}
    \centering
    \small
    \setlength{\tabcolsep}{5pt}
    \resizebox{\linewidth}{!}{
    \begin{tabular}{llccc}
        \toprule
        Model & Method & Countdown & GSM8K & SVAMP \\
        \midrule
        LLaDA-8B-Base & Baseline & 11.72 & 64.75 & 80.33 \\
        LLaDA-8B-Base & FT & 14.84 & 71.34 & 82.33 \\
        LLaDA-8B-Base & \textbf{CSRR} & \textbf{21.88} & \textbf{79.23} & \textbf{82.67} \\
        \midrule
        LLaDA-8B-Instruct & Baseline & 21.48 & 78.62 & 86.67 \\
        LLaDA-8B-Instruct & FT & 25.39 & 83.24 & 88.67 \\
        LLaDA-8B-Instruct & \textbf{CSRR} & \textbf{29.30} & \textbf{83.47} & \textbf{89.00} \\
        \bottomrule
    \end{tabular}
    }
\end{minipage}\par\medskip

\subsection{Fixed-Scale Results}
\label{sec:app-fixed-scale-results}

Table~\ref{tab:appendix-fixed-scale} reports LLaDA-8B-Base and LLaDA-8B-Instruct results when the recurrent scale is fixed to $0.5$. This setting removes the scale-selection step used in the main table and therefore provides a more constrained evaluation of the recurrent hidden-state path. CSRR remains consistently effective across most reasoning settings, indicating that the improvement is not solely an artifact of choosing the best scale separately for each benchmark configuration. The gains are especially clear for LLaDA-8B-Base and remain positive in most LLaDA-8B-Instruct settings.

\subsection{Results with Different Block Sizes}
\label{sec:app-different-blocksize-results}

\finishfigurewrap
\noindent\begin{minipage}[t]{\dimexpr0.5\textwidth-\columnsep\relax}
\vspace{0pt}
Table~\ref{tab:appendix-different-blocksize} evaluates CSRR under different semi-autoregressive block sizes with generation length $256$ and NFE $128$. Changing the block size alters how many tokens are denoised together within each local generation segment, which can affect both the baseline sampler and the usefulness of previous-step hidden states. CSRR improves over the corresponding baseline across all reported block-size settings, indicating that the method is compatible with different block granularities rather than being tuned to a single default block size.
\end{minipage}\hfill
\begin{minipage}[t]{0.5\textwidth}
\vspace{0pt}
    \captionof{table}{Reasoning results with different block sizes on LLaDA-8B-Base and LLaDA-8B-Instruct. All settings use generation length $256$ and NFE $128$. All values are percentages. Parenthesized values denote absolute improvements over the corresponding baseline.}
    \label{tab:appendix-different-blocksize}
    \centering
    \small
    \setlength{\tabcolsep}{5pt}
    \resizebox{\linewidth}{!}{
    \begin{tabular}{llccc}
        \toprule
        Block size & Method & Countdown & GSM8K & SVAMP \\
        \midrule
        \rowcolor{gray!12}
        \multicolumn{5}{l}{\textit{LLaDA-8B-Base}} \\
        \multirow{2}{*}{8} & Baseline & 7.81 & 56.18 & 71.00 \\
        & \textbf{CSRR} & 15.23 {\scriptsize(\textcolor{green!50!black}{+7.42})} & 70.66 {\scriptsize(\textcolor{green!50!black}{+14.48})} & 77.33 {\scriptsize(\textcolor{green!50!black}{+6.33})} \\
        \midrule
        \multirow{2}{*}{16} & Baseline & 8.59 & 59.36 & 74.67 \\
        & \textbf{CSRR} & 14.84 {\scriptsize(\textcolor{green!50!black}{+6.25})} & 71.27 {\scriptsize(\textcolor{green!50!black}{+11.91})} & 77.00 {\scriptsize(\textcolor{green!50!black}{+2.33})} \\
        \midrule
        \multirow{2}{*}{32} & Baseline & 10.16 & 60.88 & 77.67 \\
        & \textbf{CSRR} & 17.58 {\scriptsize(\textcolor{green!50!black}{+7.42})} & 70.66 {\scriptsize(\textcolor{green!50!black}{+9.78})} & 81.00 {\scriptsize(\textcolor{green!50!black}{+3.33})} \\
        \midrule
        \rowcolor{gray!12}
        \multicolumn{5}{l}{\textit{LLaDA-8B-Instruct}} \\
        \multirow{2}{*}{8} & Baseline & 10.94 & 73.09 & 82.00 \\
        & \textbf{CSRR} & 19.14 {\scriptsize(\textcolor{green!50!black}{+8.20})} & 77.71 {\scriptsize(\textcolor{green!50!black}{+4.62})} & 89.00 {\scriptsize(\textcolor{green!50!black}{+7.00})} \\
        \midrule
        \multirow{2}{*}{16} & Baseline & 17.19 & 75.82 & 83.67 \\
        & \textbf{CSRR} & 27.34 {\scriptsize(\textcolor{green!50!black}{+10.15})} & 77.71 {\scriptsize(\textcolor{green!50!black}{+1.89})} & 86.33 {\scriptsize(\textcolor{green!50!black}{+2.66})} \\
        \midrule
        \multirow{2}{*}{32} & Baseline & 13.67 & 76.04 & 85.67 \\
        & \textbf{CSRR} & 24.22 {\scriptsize(\textcolor{green!50!black}{+10.55})} & 78.54 {\scriptsize(\textcolor{green!50!black}{+2.50})} & 86.00 {\scriptsize(\textcolor{green!50!black}{+0.33})} \\
        \bottomrule
    \end{tabular}
    }
\end{minipage}\par\medskip

\noindent\begin{minipage}{\linewidth}
\subsection{Source and Injection Layer Positions}
\label{sec:app-source-injection-positions}
\noindent\begin{minipage}[t]{\dimexpr0.5\linewidth-\columnsep\relax}
\vspace{0pt}
Figure~\ref{fig:source-injection-position-heatmap} studies how the choice of source layer and injection layer affects CSRR. The source layer denotes the transformer block output used as the previous-step hidden state, and the injection layer denotes the block into which this state is incorporated during the current denoising step. The default CSRR configuration extracts the hidden state after the final block and injects it inside the first block. We conduct this sweep using LLaDA-8B-Base with generation length $256$, NFE $128$, and recurrent scale $1.0$. The results show that most evaluated source-injection pairs improve over the corresponding baseline, supporting the robustness of the recurrent hidden-state path across different layer choices. The default pair remains among the stronger settings, while several alternative pairs achieve even higher accuracy, suggesting that layer-position selection is a meaningful design dimension that deserves further study.
\end{minipage}\hfill
\begin{minipage}[t]{0.5\linewidth}
\vspace{0pt}
    \centering
    \includegraphics[width=\linewidth]{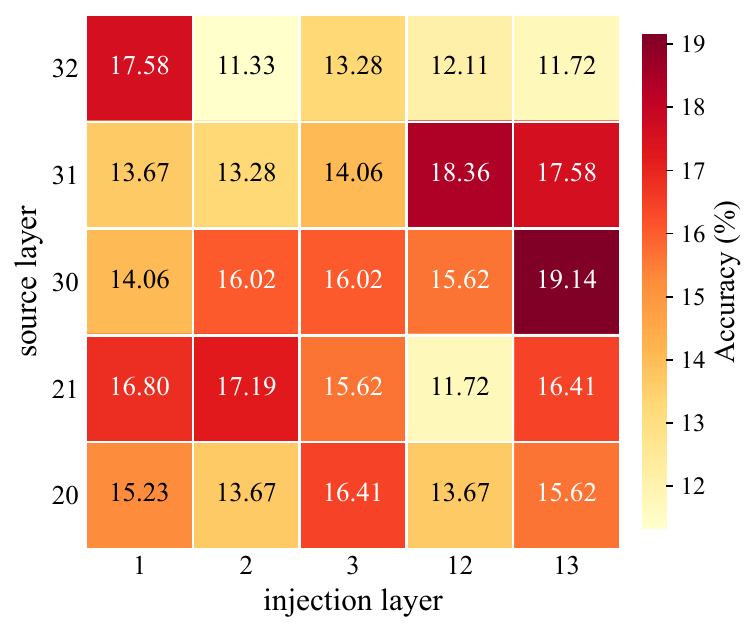}
    \captionof{figure}{Accuracy under different source-layer and injection-layer choices for CSRR. The sweep uses LLaDA-8B-Base with generation length $256$, NFE $128$, and recurrent scale $1.0$.}
    \label{fig:source-injection-position-heatmap}
\end{minipage}
\end{minipage}\par

\section{Evaluation Prompts}
\label{sec:evaluation-prompts}

This section reports the prompt templates used in evaluation. The templates are shown with placeholders for instance-specific fields, and line breaks are inserted for readability. The actual input sequence is obtained by applying the corresponding model tokenizer's chat template to the user message, followed by an optional assistant-side prefill for reasoning tasks.

\subsection{Chat-Template Wrapping}
\label{sec:chat-template-wrapping}

For LLaDA-8B-Base and LLaDA-8B-Instruct, the tokenizer wraps each user message in the following structure:
\begingroup
\footnotesize
\begin{verbatim}
<|startoftext|>
<|start_header_id|>user<|end_header_id|>

{user message}<|eot_id|>
<|start_header_id|>assistant<|end_header_id|>

{optional assistant prefill}
\end{verbatim}
\endgroup
For Dream-7B-Base and Dream-7B-Instruct, the tokenizer additionally inserts the default system message used by the model tokenizer:
\begingroup
\footnotesize
\begin{verbatim}
<|im_start|>system
You are a helpful assistant.<|im_end|>
<|im_start|>user
{user message}<|im_end|>
<|im_start|>assistant
{optional assistant prefill}
\end{verbatim}
\endgroup
For Countdown, GSM8K, and SVAMP, the optional assistant prefill is \texttt{<reasoning>}. For HumanEval, HumanEval+, and MBPP, no assistant prefill is used.

\subsection{Reasoning Prompts}
\label{sec:reasoning-prompts}

The reasoning benchmarks use task-specific instructions and require the final answer to be wrapped in \texttt{\textbackslash boxed\{\}}. Countdown uses the arithmetic-expression prompt below:
\begingroup
\footnotesize
\begin{verbatim}
Using only the provided numbers, create an
arithmetic expression that evaluates to 
exactly the provided target
number. You may use the operations +, -, *,
and / as needed, but each number must be 
used exactly once. Think step-by-step.
After reasoning, provide only your final
expression inside \boxed{} tags without 
including an equals sign or the target 
number. For example: \boxed{a + b * c}
Respond in the following format:
<reasoning>
Your reasoning here
</reasoning>
<answer>
\boxed{...}
</answer>

Numbers: {numbers}
Target: {target}
\end{verbatim}
\endgroup
GSM8K uses the following mathematical-reasoning prompt:
\begingroup
\footnotesize
\begin{verbatim}
You are a math expert. You will be given a 
question to solve. Solve it step by step. 
Wrap the final answer in a \boxed{}.
Respond in the following format:
<reasoning>
Your reasoning here
</reasoning>
<answer>
\boxed{...}
</answer>

{question}
\end{verbatim}
\endgroup
SVAMP uses the same response format as the mathematical-reasoning prompt, with the problem text formed from the dataset body and question fields:
\begingroup
\footnotesize
\begin{verbatim}
You are a math expert. You will be given a 
question to solve. Solve it step by step. 
Wrap the final answer in a \boxed{}.
Respond in the following format:
<reasoning>
Your reasoning here
</reasoning>
<answer>
\boxed{...}
</answer>

{body}{question}
\end{verbatim}
\endgroup

\subsection{Code-Generation Prompts}
\label{sec:code-generation-prompts}

HumanEval and HumanEval+ share the same prompt template, where \texttt{\{function\_prompt\}} contains the function signature and docstring from the benchmark instance:
\begingroup
\footnotesize
\begin{verbatim}
Complete the following Python function:
{function_prompt}
\end{verbatim}
\endgroup
MBPP uses a task prompt together with the provided unit tests:
\begingroup
\footnotesize
\begin{verbatim}
You are a coding expert. You will be given a
coding problem to solve. Write a Python 
function to solve the problem. Output only
the function code wrapped in ```python``` 
blocks. Do not include explanations or extra 
text.

{problem}

Your code should pass the following tests:

{unit_tests}
\end{verbatim}
\endgroup

\end{document}